\documentclass{article}

 \usepackage[preprint]{neurips_2026}

\usepackage[utf8]{inputenc} %
\usepackage[T1]{fontenc}    %
\usepackage{hyperref}       %
\usepackage{url}            %
\usepackage{booktabs}       %
\usepackage{amsfonts}       %
\usepackage{nicefrac}       %
\usepackage{microtype}      %
\usepackage{xcolor}         %
\usepackage{graphicx}
\usepackage{wrapfig}
\usepackage{amsmath}

\usepackage{framed}

\usepackage{caption}

\definecolor{formalshade}{rgb}{0.95,0.95,0.97}
\definecolor{darkblue}{rgb}{0.14,0.22,0.52}

\newenvironment{formal}{
  
  \MakeFramed{\advance\hsize-\width\FrameRestore}
  \noindent\hspace{-4.55pt}
}{
  \endMakeFramed
}

\title{PlatonicNav: Unveiling Semantic Correspondence in Navigation with Platonic Topological Maps}

\author{
Junlin Long$^{1*}$\quad
Zeyu Zhang$^{2*\dag}$\quad
Xu Deng$^{3*}$\quad
Yiran Wang$^{1*}$\\
\textbf{Yue Yang}$^{2}$\quad
\textbf{Luke Borgnolo}$^{2}$\quad
\textbf{Maxwell Twelftree}$^{2}$\quad
\textbf{Yang Zhao}$^{4\ddag}$\\
[0.3em]
$^1$USYD\quad
$^2$Maincode\quad
$^3$UNSW\quad
$^4$La Trobe\\
[0.1em]
\footnotesize $^*$Equal contribution.
$^\dag$Project lead.
$^\ddag$Corresponding author: y.zhao2@latrobe.edu.au.
}

\begin{document}

\maketitle

\begin{abstract}
Embodied visual navigation, where an agent perceives a complex environment and acts to reach a goal from raw sensory input, underpins a wide range of applications such as household service robotics, assistive robotics, and large-scale autonomous exploration. However, recent attempts to unify vision-and-language navigation (VLN) and object goal navigation (ObjNav) remain at the level of architectural fusion, mixed-task training, and large vision-language pretraining, without examining whether independently trained vision and language encoders may already share a common semantic structure. Moreover, even object-centric topological maps still ground language goals through explicit cross-modal supervision such as CLIP or large vision-language models, leaving open whether such grounding is possible from a purely vision-built map. To address these challenges, we extend the \emph{Platonic Representation Hypothesis} to embodied navigation and recast vision-only ObjNav, cross-modal ObjNav, and VLN as three different interfaces to the same object-centric semantic manifold. We further introduce \textbf{PlatonicNav}, a training-free framework whose \textbf{Platonic Topological Map} fuses geometric and semantic node distances from a self-supervised visual encoder, and grounds language goals via \emph{blind matching} without any paired vision-language data. Extensive experiments on simulation benchmarks including HM3D-IIN, OVON, and R2R-CE on MP3D, together with deployment on Unitree Go2, which demonstrate that \textbf{PlatonicNav} generalizes across tasks, modalities, and embodiments without explicit cross-modal training.
Code:~\url{https://github.com/AIGeeksGroup/PlatonicNav}.
Website:~\url{https://aigeeksgroup.github.io/PlatonicNav}.
\end{abstract}

\section{Introduction}

Embodied visual navigation, where an agent perceives a complex environment and acts to reach a goal from raw sensory input, underpins a wide range of applications, including household service robots, assistive robotics, autonomous exploration in unknown environments, and augmented reality systems. Two representative paradigms have shaped almost all modern progress: Vision-and-Language Navigation (VLN)~\cite{krantz_vlnce_2020}, in which an agent follows natural-language instructions grounded in visual observations, and Object Goal Navigation (ObjNav), in which the agent must locate a target object specified by a semantic category. They are typically studied as distinct problems, with VLN emphasizing multimodal reasoning and long-horizon instruction following, and ObjNav emphasizing semantic understanding and goal-directed exploration. Yet behind these different \emph{interfaces}, both tasks ask the same agent to connect visual observations, object-level semantics, and spatial decisions over the same physical scene, hinting at a shared structural foundation that the field has not yet made explicit.

Two challenges stand in the way of making this foundation explicit. \emph{First}, although a growing body of work~\cite{liu2025nav,zhang2024uni,gao2025octonav} attempts to unify VLN and ObjNav within a single navigation foundation model, these efforts remain at the level of architectural fusion, mixed-task training, and large vision-language pretraining; they ask how to engineer a single model that handles both, while leaving open whether independently trained vision and language encoders may \emph{already} share a common semantic structure, so that the field would be paying for cross-modal supervision that is, in part, redundant. \emph{Second}, even when navigation systems adopt object-centric topological maps~\cite{garg2024robohop,garg2025objectreact}, which by construction sit close to the underlying semantic structure of the scene, language goals are still grounded into them through explicit cross-modal supervision such as CLIP~\cite{radford2021learning} or large vision-language models~\cite{bai2025qwen3}, and vision-only ObjNav, cross-modal ObjNav, and VLN remain three disjoint task interfaces, even though a single semantic structure is already accessible from a purely vision-built map.

\begin{wrapfigure}{r}{0.5\linewidth}
    \centering
    
    \includegraphics[width=\linewidth]{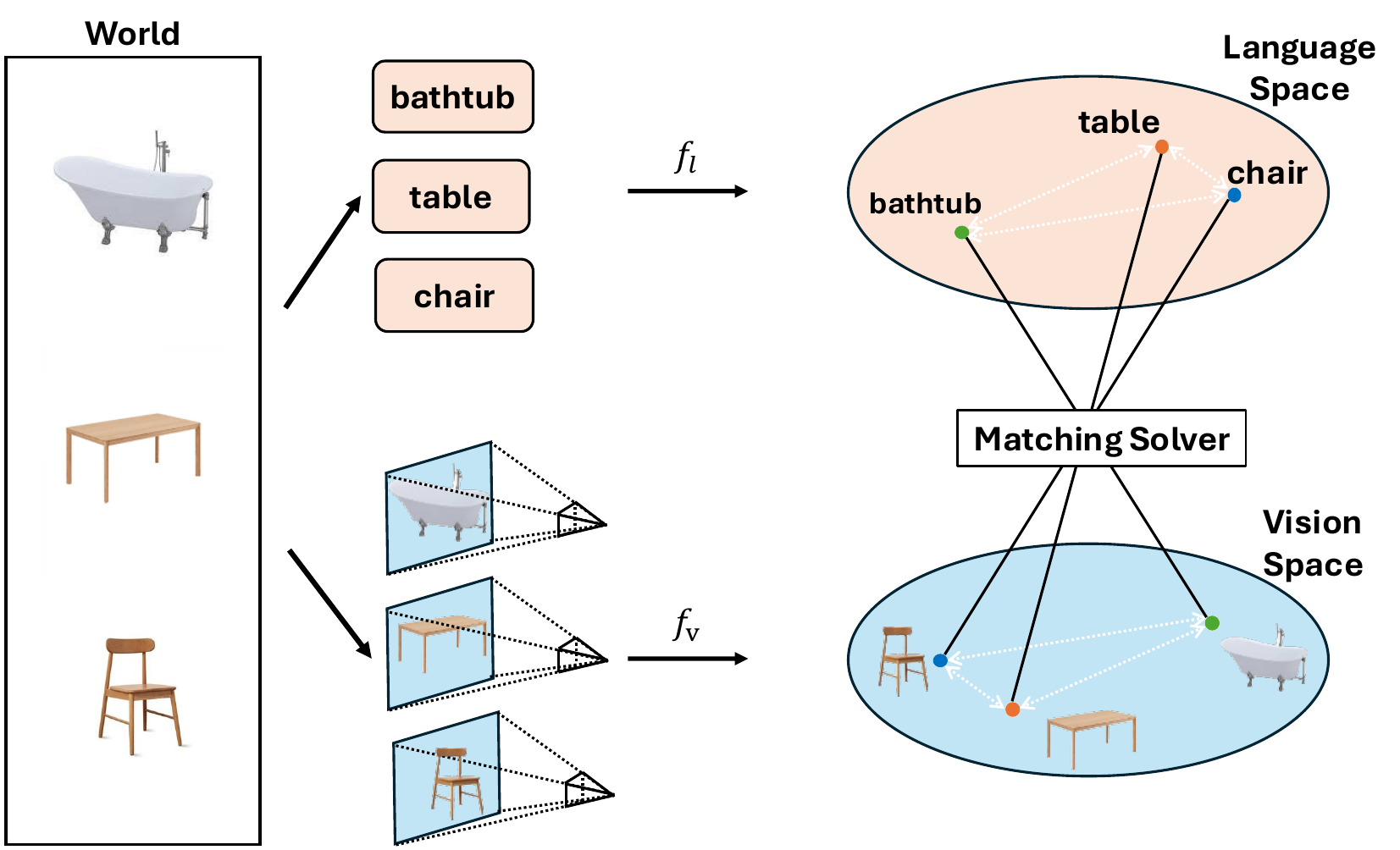}
    
    \caption{
    \textbf{Blind matching of vision and language in navigation scene.}
    Text and images are both abstractions of the same underlying world. Vision and language encoders $f_v$ and $f_l$ learn similar pairwise relations between concepts. We exploit these pairwise relations in a matching solver to recover valid correspondences between vision and language representations without requiring any paired data~\cite{schnaus2025s}.
    }
   
    \label{fig:blind-match}
\end{wrapfigure}

Two recent observations point to a way out. \emph{First}, the \emph{Platonic Representation Hypothesis}~\cite{huh2024platonic} argues that models trained on different modalities and objectives converge toward a shared statistical model of reality in their representation spaces, so that visual and language semantic distances become implicitly aligned even when the encoders are never exposed to paired data. If this property carries over from static representation learning to embodied navigation, the natural prediction is that vision-only ObjNav, cross-modal ObjNav, and VLN should produce closely related trajectories in the same scene with the same target, since all three are then simply querying the same underlying semantic geometry through different goal interfaces. \emph{Second}, an object-centric topological map~\cite{garg2024robohop,garg2025objectreact} is itself a discrete approximation of that geometry: its nodes are object segments produced by a self-supervised visual encoder~\cite{simeoni2025dinov3}, and pairwise node distances already encode visual semantics. This makes the map a natural substrate for connecting a language goal to a visual node directly through the relational structure of the two encoders, via \emph{blind matching}~\cite{schnaus2025s}, without any paired vision-language data, contrastive pretraining, or VLM supervision. On this view, the cross-modal alignment that current systems engineer is, in significant part, recovered for free from geometry that already exists in the representations.

Building on these observations, we propose \textbf{PlatonicNav}, a training-free framework that grounds language goals into a vision-built \textbf{Platonic Topological Map} via \emph{blind matching}, casting vision-only ObjNav, cross-modal ObjNav, and VLN as three instances of navigation over a single object-centric semantic manifold. The contributions of this paper are summarized as follows:
\begin{itemize}
    \item We formulate embodied navigation through the lens of the \emph{Platonic Representation Hypothesis}~\cite{huh2024platonic} and propose a falsifiable two-step thought experiment that turns the representation-level unification of vision-only ObjNav, cross-modal ObjNav, and VLN into a testable claim on real navigation trajectories.
    \item We introduce \textbf{PlatonicNav}, a training-free framework built on \textbf{Platonic Topological Maps} whose edges fuse geometric and Platonic semantic distances, with language goals grounded into the map via \emph{blind matching}~\cite{schnaus2025s} between independently trained vision~\cite{simeoni2025dinov3} and language~\cite{ni-etal-2022-large} encoders, requiring no paired vision-language data.
    \item We evaluate \textbf{PlatonicNav} on simulation benchmarks (HM3D-IIN, OVON, and R2R-CE on MP3D) and on real-world quadruped platforms (Unitree Go2), showing cross-task, cross-modality, and cross-embodiment generalization without explicit cross-modal training.
\end{itemize}

Together, these results suggest that the cross-modal alignment that today's navigation systems engineer with paired data, contrastive learning, or VLM supervision is, in significant part, already latent in independently trained vision and language encoders, and that ObjNav and VLN can be understood as different interfaces to the same object-centric semantic structure of the environment. We view this as a first step toward a representation-centric view of embodied navigation, where map and policy are organized around the geometry of meaning, in a regime where the underlying representations no longer respect modality boundaries.

\section{Related Work}

\subsection{Embodied Visual Navigation}

Embodied visual navigation has crystallized around two long-running task families. Object Goal Navigation traces back to imitation- and RL-based agents trained inside Habitat~\cite{savva2019habitat}: DD-PPO~\cite{wijmans2020ddppo} pushed RL to billions of frames, Habitat-Web~\cite{ramrakhya2022habitatweb} and PIRLNav~\cite{ramrakhya2023pirlnav} grafted human demonstrations and DAgger-style~\cite{ross2011dagger} fine-tuning onto policy training, and SemExp~\cite{chaplot2020semexp} introduced explicit semantic maps. As models grew, the field pivoted to \emph{zero-shot} formulations that route through foundation models: ZSON~\cite{majumdar2022zson} repurposes multimodal goal embeddings, CoWs~\cite{gadre2023cows} and L3MVN~\cite{yu2023l3mvn} let CLIP and LLMs steer frontier exploration, ESC~\cite{zhou2023esc} adds soft commonsense priors, LFG~\cite{shah2023lfg} treats LLMs as semantic guesswork heuristics, and VLM-grounded mappers such as VLFM~\cite{yokoyama2024vlfm} and VLMaps~\cite{huang2023vlmaps} write language-aligned features into 2D occupancy. Open-vocabulary benchmarks~\cite{yokoyama2024hm3dovon} have since become the default stress test. Vision-and-Language Navigation evolved on a parallel track: from R2R~\cite{anderson2018r2r} and its multilingual successor RxR~\cite{ku2020rxr} through the continuous-control reformulation VLN-CE~\cite{krantz_vlnce_2020}, with method advances spanning history-aware transformers~\cite{chen2021hamt}, dual-scale graph reasoning~\cite{chen2022duet}, BEV pretraining~\cite{an2023bevbert}, and topological-graph waypoint prediction in ETPNav~\cite{an2024etpnav}; more recent VLM-as-policy work, including NaVid~\cite{zhang2024navid}, NaVILA~\cite{cheng2024navila}, NavGPT~\cite{zhou2024navgpt}, and MapGPT~\cite{chen2024mapgpt}, collapses the perception-planning stack into a single video-conditioned generator. A nascent unification thread argues that ObjNav and VLN should share one backbone: Uni-NaVid~\cite{zhang2024uni}, OctoNav~\cite{gao2025octonav}, Nav-R1~\cite{liu2025nav}, MobileVLA-R1~\cite{huang2025mobilevla}, and MTU3D~\cite{zhu2025mtu3d} pursue this through mixed-task training, large-scale vision-language pretraining, or 3D-grounded reasoning, with unification enacted at the level of architectures and data mixtures.

\subsection{Representation-Level Grounding for Navigation}

A second axis of the literature concerns how language is bound to vision. The dominant paradigm remains \emph{paired supervision}: CLIP~\cite{radford2021learning}, ALIGN~\cite{jia2021align}, and SigLIP~\cite{zhai2023siglip} engineer a joint embedding from massive image-text corpora, and recent vision-language models such as Qwen3-VL~\cite{bai2025qwen3} extend this lineage to instruction-tuned multimodal generation; downstream, the paradigm reappears in open-vocabulary 3D representations such as OpenScene~\cite{peng2023openscene}, ConceptFusion~\cite{jatavallabhula2023conceptfusion}, and ConceptGraphs~\cite{gu2024conceptgraphs}, which lift CLIP-style features into voxels, point clouds, or scene graphs. A separate lineage prioritizes graph sparsity over dense metric reconstruction, beginning with semi-parametric memory~\cite{savinov2018sptm} and Neural Topological SLAM~\cite{chaplot2020nts}, and culminating in the segment-based formulation of RoboHop~\cite{garg2024robohop} and the object-relative pipeline of ObjectReact~\cite{garg2025objectreact}. Cutting across both threads, the \emph{Platonic Representation Hypothesis}~\cite{huh2024platonic} posits that encoders trained on disjoint modalities converge toward a common representation of reality; subsequent work has both refined and contested this view: an Aristotelian critique~\cite{groger2026revisiting} questions naive isotropy, large-scale re-examinations~\cite{koepke2026back} probe its scaling behavior, and JAM~\cite{yoon2025jam} argues that residual misalignment can be closed post-hoc. Operationally, this convergence is exploited by \emph{blind matching}~\cite{schnaus2025s}, which recovers vision-language correspondence from pairwise relational structure alone, dispensing with parallel data. Underwriting both encoder sides are self-supervised vision learners such as DINO~\cite{caron2021dino}, DINOv2~\cite{oquab2024dinov2}, DINOv3~\cite{simeoni2025dinov3}, and MAE~\cite{he2022mae}, alongside dense text retrievers such as Sentence-BERT~\cite{reimers2019sbert}, GTR~\cite{ni-etal-2022-large}, and T5~\cite{raffel2020exploring}.

\begin{figure}[t]
    \centering
    \includegraphics[width=\linewidth]{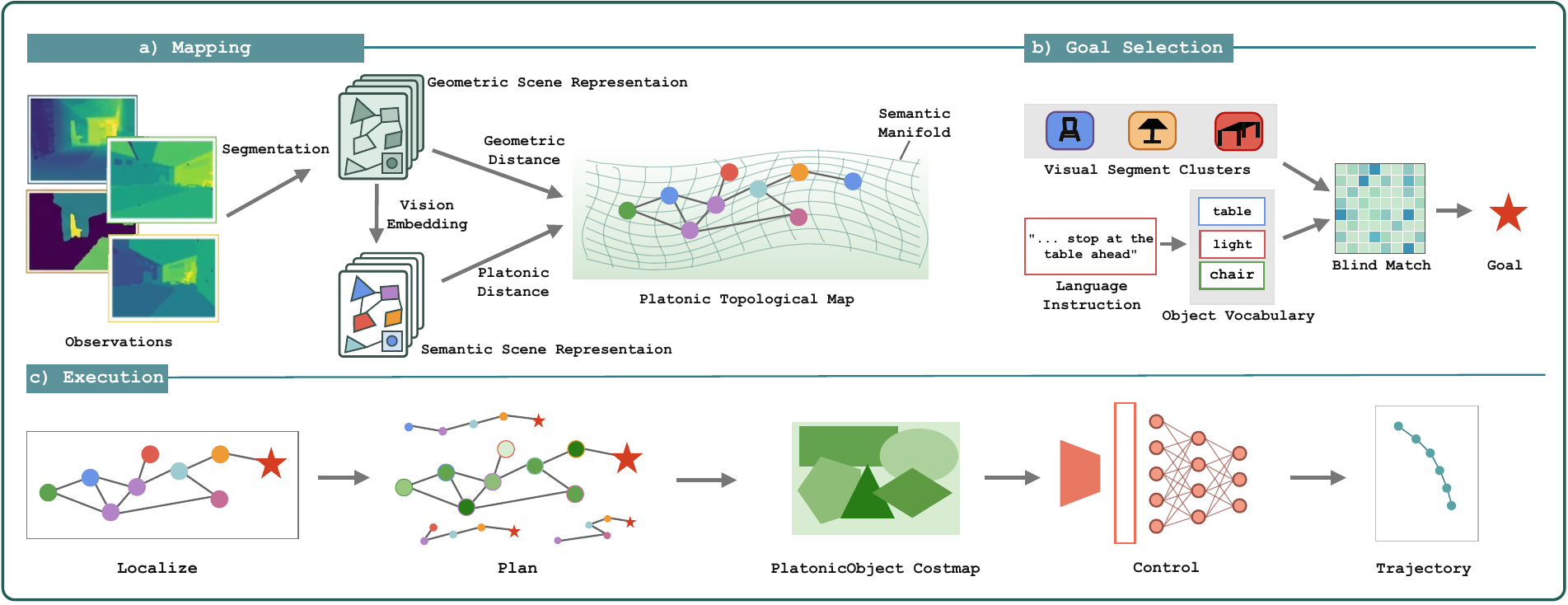}
    \caption{
    \textbf{PlatonicNav Pipeline.}
    \textbf{(a) Mapping:} We construct Platonic Topological Map as a semantic scene graph, where image segments are used as object nodes, and edges are weighted by both geometric distance and semantic distance computed from vision embedding space.
    \textbf{(b) Goal Selection:} Given the natural-language instruction, we pairwise blind match language embeddings of goal object category and visual embedding of segment cluster, selecting the candidate goal nodes in Platonic Topological Map.
    \textbf{(c) Execution:} Given the map and candidate goal nodes, we compute the paths to the goal node which can be reached by lightest edge weight; the resulting path lengths are assigned to segmentation masks to form a \emph{PlatonicObject Costmap} for control prediction. }
    \label{fig:top-pipeline}
\end{figure}

\section{Method}
\label{sec: method}

\subsection{Overview}
PlatonicNav consists of three components. (i) A representation-level framing of embodied navigation that recasts vision-only ObjNav, cross-modal ObjNav, and VLN as three goal interfaces over the same object-centric semantic manifold (Section~\ref{sec:platonic_in_nav}). (ii) A \emph{Platonic Topological Map} that enriches segment-based topological graphs (for segment-based topological graph, see~\ref{sec:topo-map-for-em} with relational semantic distances over a self-supervised visual encoder, and grounds language queries into the map via blind matching, requiring no paired vision-language data (Section~\ref{sec:platonic_topomap})(for Platonic Representation Hypothesis, see~\ref{sec: prelim-platonic}. (iii) A real-world deployment on a Unitree Go2 Air quadruped that validates the framework beyond simulation (Section~\ref{sec:real_world}).

\subsection{Platonic Representation Hypothesis in Embodied Navigation}
\label{sec:platonic_in_nav}

We extend the \emph{Platonic Representation Hypothesis} to embodied navigation and argue that vision-only and language-conditioned navigation share an underlying semantic structure that can be exploited without explicit cross-modal supervision.

\paragraph{Setup.}
Let $f_v: \mathcal{I} \rightarrow \mathbb{R}^{d_v}$ be a visual encoder trained with self-supervised objectives (e.g., DINOv3~\cite{simeoni2025dinov3}), and $f_l: \mathcal{T} \rightarrow \mathbb{R}^{d_l}$ an independently pretrained language encoder (e.g., GTR-T5~\cite{ni-etal-2022-large}); the two embedding spaces in general have different dimensionalities and unrelated coordinate frames. Given any finite set of concepts $\{x_i\}$, each realized as a visual exemplar $x_i^{\text{img}}$ and a textual description $x_i^{\text{txt}}$, we form the pairwise distance matrices
\begin{equation}
    D^v_{ij} = \|f_v(x_i^{\text{img}}) - f_v(x_j^{\text{img}})\|, \qquad D^l_{ij} = \|f_l(x_i^{\text{txt}}) - f_l(x_j^{\text{txt}})\|.
\end{equation}
We instantiate the relational transform $\mathcal{N}(\cdot)$ in Eq.~\eqref{eq:platonic-alignment} (Appendix~\ref{sec: prelim-platonic}) by double-centering,
\begin{equation}
    \mathcal{N}(D)_{ij} \;=\; D_{ij} - \bar{D}_{i\cdot} - \bar{D}_{\cdot j} + \bar{D}_{\cdot\cdot},
\end{equation}
which removes modality-specific row, column, and global means and isolates each concept's relational position. Under this choice, the alignment $\mathcal{N}(D^v) \approx \mathcal{N}(D^l)$ asserts that semantic relationships (similarity, hierarchy) are preserved across modalities even when $f_v$ and $f_l$ never share training data.

\paragraph{Implication for vision-only ObjNav.}
In vision-only topological-map-based navigation, the environment is represented as a graph $\mathcal{G} = (\mathcal{V}, \mathcal{E})$, where each node $v_i \in \mathcal{V}$ corresponds to an object-centric image segment $s_i$ with visual embedding $\mathbf{z}_i = f_v(s_i) \in \mathbb{R}^{d_v}$. To exploit Eq.~\eqref{eq:platonic-alignment} for navigation, we make the embedding geometry a principal driver of edge weights:
\begin{equation}
    d_{\mathcal{G}}(v_i, v_j) \;:=\; \|\mathbf{z}_i - \mathbf{z}_j\|,
\end{equation}
so that path lengths over $\mathcal{G}$ approximate geodesic distances on the semantic manifold induced by $f_v$. Under this design choice, planning over $\mathcal{G}$ reduces to finding a trajectory that minimizes semantic discrepancy to the goal. Section~\ref{sec:platonic_topomap} refines $d_{\mathcal{G}}$ into a hybrid metric that fuses geometric and semantic distances.

\paragraph{Connection to cross-modal ObjNav and VLN.}
In both cross-modal ObjNav and VLN, the goal is specified in the language space. Under the Platonic Representation Hypothesis, the language query embedding $\mathbf{u} = f_l(t)$ and the corresponding visual node embedding $\mathbf{z}_i$ are not assumed to coincide in absolute coordinates; their alignment lives in the relational structure each induces with neighboring concepts. Identifying the goal node thus reduces to finding the visual node whose pairwise relations to other nodes match the language-side pairwise relations of the query, a relational match we formalize via blind matching in Section~\ref{sec:platonic_topomap}. Cross-modal ObjNav and VLN can therefore be interpreted as navigating on the same semantic manifold as vision-only ObjNav, with the query supplied through language.

\paragraph{Unifying Perspective.}
This perspective reveals that vision-only ObjNav, cross-modal ObjNav, and VLN are three goal interfaces over the same shared semantic manifold, distinguished by how the goal is presented. For example, when searching for a ``cup'', an agent implicitly leverages semantic priors: a cup is more likely to be found in a kitchen or a living room than on a bed in a bedroom. Such reasoning emerges from the structure of the representation space, where objects co-occurring in similar contexts are embedded closer together. We therefore propose to unify VLN and ObjNav at the \emph{representation level} by exploiting the shared semantic geometry of independently trained visual and language embeddings, leading to a principled framework for \emph{Platonic Topological Maps} and language-to-map grounding via blind matching.

\begin{figure}[t]
    \centering
    \includegraphics[width=\linewidth]{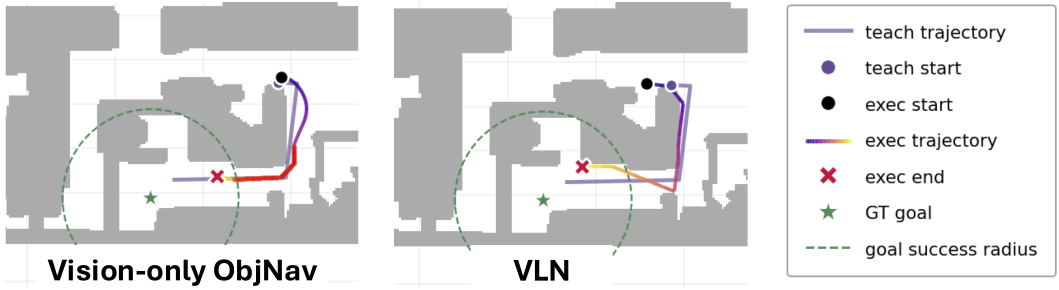}
    \caption{
    \textbf{Visual-only ObjNav, VLN, and PlatonicNav trajectory comparison.}
    Top-down trajectory maps of vision-only ObjNav (ObjectReact), VLN (ETPNav), and PlatonicNav with matched scenes and targets, corresponding to Step 1 and Step 2 of our thought experiment (Section~3.1). Trajectory similarity suggests that vision-only navigation implicitly encodes language-level semantic structure, motivating our investigation of the Platonic Representation Hypothesis~\cite{huh2024platonic} in embodied navigation.
    }
    \label{fig:language-vision-align}
\end{figure}

\paragraph{A testable thought experiment.}
The above hypothesis leads to a two-step thought experiment. Throughout, each trajectory $\tau = (v_1, \dots, v_T)$ denotes a sequence of visited object-centric nodes in the topological map, and trajectory similarity is measured via overlap of visited-node sets, semantic scene categories, and landmark intersections, rather than pixel-level path comparison.

\emph{Step~1: Motivation.} We place a vision-only ObjNav agent (ObjectReact) and a jointly trained vision-language VLN agent (ETPNav~\cite{an2024etpnav}) in the same scene with the same target object instance, with the goal specified to ObjectReact through its segmentation mask and to ETPNav through the corresponding language description. Since the original ObjectReact and ETPNav checkpoints are released on different scene sets (HM3D and MP3D respectively), we port ObjectReact to MP3D to enable a same-scene comparison. If
\begin{equation}
\label{eq:step1}
    \tau^{\text{vision-only ObjNav}} \approx \tau^{\text{VLN}},
\end{equation}
this suggests that pure-vision navigation already encodes language-level semantic structure (see Fig.~\ref{fig:language-vision-align}). However, Eq.~\eqref{eq:step1} alone is insufficient: trajectory overlap may partly reflect environmental constraints (e.g., floor-plan geometry), and the cross-modal target correspondence relies on human annotation, which constitutes explicit cross-modal supervision since the target identity is provided by the dataset rather than discovered by the representation space.

\emph{Step~2: Critical test.} We construct a PlatonicNav agent that grounds a language goal into the visual topological map via blind matching (cf.\ Section~\ref{sec:platonic_topomap}, Eq.~\eqref{eq:blind_matching}): visual cluster centroids $\{\bar{\mathbf{z}}_k\}$ are matched to language category embeddings $\{\mathbf{u}_k\}$ using relational structure alone, without any paired vision-language supervision. We compare PlatonicNav with cross-modal ObjNav baselines that ground goals through contrastive-trained models such as VLMs on HM3D-OVON. The OVON comparison (see Tab.~\ref{tab:hm3d-ovon}) further strengthens our proposition: even though PlatonicNav uses no explicit cross-modal training, its non-trivial navigation performance indicates that the relational geometry shared by independently trained vision and language representations is sufficiently preserved to support embodied goal grounding. Beyond the trajectory-level intuition motivated in Step~1, this result provides direct evidence that the \emph{Platonic Representation Hypothesis} carries over to embodied navigation.

\subsection{Platonic Topological Maps}
\label{sec:platonic_topomap}

We introduce \emph{Platonic Topological Maps}, a representation-centric extension of segment-based topological maps that explicitly exploits the Platonic Representation Hypothesis in embodied navigation.

\paragraph{From Topological Maps to Platonic Topological Maps.}
We build upon ObjectReact~\cite{garg2025objectreact}, a representative vision-only object-centric topological navigation framework, where the environment is modeled as a topometric graph. Each node corresponds to an object segment extracted from visual observations, and edges encode spatial or associative relationships (cf. Fig.~\ref{fig:top-pipeline}). In standard ObjectReact, node connectivity and path planning are primarily governed by 3D geometric proximity and object-association heuristics; topology is treated as a \emph{purely geometric construct}, ignoring the underlying \emph{semantic structure} of the representation space.

\paragraph{Key idea.}
Our central insight is that a topological map should not be defined purely over physical space, but over a \emph{semantic manifold} $\mathcal{M}$ induced by self-supervised visual representations (Fig.~\ref{fig:top-pipeline}). We reinterpret the topological graph as a \emph{Platonic graph}, in which node-to-node distances reflect semantic proximity in $\mathbb{R}^{d_v}$ in addition to geometric distance, and view the resulting graph as a discretization of $\mathcal{M}$ over which navigation is geodesic traversal under a learned metric.

\paragraph{Node representation.}
Each object segment $s_i$ is associated with a visual embedding
\begin{equation}
    \mathbf{z}_i \;=\; f_v(s_i) \;\in\; \mathbb{R}^{d_v},
\end{equation}
where $f_v$ is a self-supervised visual encoder (e.g., DINOv3~\cite{simeoni2025dinov3}). Following RoboHop~\cite{garg2024robohop}, we obtain $\mathbf{z}_i$ by passing the full image through $f_v$ and mean-pooling the resulting patch tokens within the mask of $s_i$, then $\ell_2$-normalizing the result so that Euclidean and cosine distances on $\{\mathbf{z}_i\}$ are rank-equivalent. Under the Platonic Representation Hypothesis, $\mathbf{z}_i$ captures semantic relationships aligned with language representations, even without explicit visual-language supervision.

\paragraph{Platonic distance.}
We define the \emph{Platonic distance} between two nodes as the cosine distance between L2-normalized visual embeddings (selected by ablation study as shown in Appendix~\ref{sec:raw_d_plat_ablation}):
\begin{equation}
\label{eq:d_plat}
    d_{\mathrm{plat}}(i,j) \;=\; 1 - \frac{\mathbf{z}_i^\top \mathbf{z}_j}{\|\mathbf{z}_i\|_2 \, \|\mathbf{z}_j\|_2}.
\end{equation}
This distance encodes semantic similarity between object segments, such that semantically related objects (e.g., \textit{chair} and \textit{table}) lie closer in the embedding space.

\paragraph{Hybrid edge weight.}
The geometric distance $d_{\text{geo}}(i,j)$ is the 3D Euclidean distance between segment centroids reconstructed from monocular depth, as in ObjectReact~\cite{garg2025objectreact}. To combine geometry and semantics on a common scale, we keep $d_{\text{geo}}$ on the same meter scale as ObjectReact's~\cite{garg2025objectreact} original geometric edge cost, colibrate $d_{\mathrm{plat}}$ to the same meter scale as $d_{\text{geo}}$ (See Appendix~\ref{sec:d_plat_meter} Eq.~\eqref{eq:d_plat_meter}), denote the normalized quantity as $\tilde{d}_{\mathrm{plat}}$, and define the edge weight as their convex combination:
\begin{equation}
\label{eq:hybrid_edge_weight}
    d(i,j) \;=\; \lambda_g \, d_{\text{geo}}(i,j) + \lambda_s \, \tilde{d}_{\mathrm{plat}}(i,j), \qquad \lambda_g, \lambda_s \geq 0, \quad \lambda_g + \lambda_s = 1.
\end{equation}
Setting $\lambda_g = 1$ recovers ObjectReact's purely geometric edge weights; $\lambda_s = 1$ ignores geometry; intermediate values trade off between the two. We treat $(\lambda_g, \lambda_s)$ as hyperparameters whose values are reported in Section~\ref{sec:experiments} and found by ablation study in Appendix~\ref{sec:lambda_ablation}.

\paragraph{Bridging the cross-space and granularity gaps.}
Two issues block a direct distance computation $\|\mathbf{z}_i - \mathbf{u}\|$ between a visual node embedding $\mathbf{z}_i$ and a language category embedding $\mathbf{u} = f_l(c)$. \emph{Fundamentally}, $f_v$ and $f_l$ map inputs into different embedding spaces, with potentially different dimensionalities $d_v \neq d_l$ and unrelated coordinate frames; absolute distances across modalities are therefore ill-defined regardless of granularity. \emph{Operationally}, even when comparison is restricted to relational structure (as blind matching does), the language vocabulary is category-level (e.g., ``chair'', ``table'') while visual segments are instance-level, so a direct relational match still requires bridging this granularity. Both observations motivate a two-stage construction: we work over relational structure to bypass the cross-space issue, and cluster visual nodes into category-level prototypes via $K$-means to bridge the granularity issue.

Concretely, given $N$ visual node embeddings $\{\mathbf{z}_i\}_{i=1}^{N}$ and a closed vocabulary of $K$ object categories $\{c_k\}_{k=1}^{K}$, we run $K$-means on $\{\mathbf{z}_i\}$ to obtain a cluster assignment $\sigma: [N] \to [K]$ and visual cluster centroids
\begin{equation}
    \bar{\mathbf{z}}_k \;=\; \frac{1}{|S_k|} \sum_{i \in S_k} \mathbf{z}_i, \qquad S_k = \{i : \sigma(i) = k\}.
\end{equation}
On the language side, the corresponding category embeddings are $\mathbf{u}_k = f_l(c_k)$, produced by an independently pretrained language encoder (e.g., GTR-T5~\cite{ni-etal-2022-large}). Both modalities now expose $K$ comparable units.

\paragraph{Goal visual cluster grounding via blind matching.}
We recover a bijection $\pi^\star \in \mathcal{S}_K$ that minimizes pairwise relational distortion between the visual and language similarity structures, following the quadratic assignment formulation of~\cite{schnaus2025s}:
\begin{equation}
\label{eq:blind_matching}
    \pi^\star \;=\; \arg\min_{\pi \in \mathcal{S}_K} \sum_{k,l=1}^{K} \big( D^v_{kl} - D^l_{\pi(k), \pi(l)} \big)^2,
\end{equation}
where $D^v_{kl} = \|\bar{\mathbf{z}}_k - \bar{\mathbf{z}}_l\|$, $D^l_{kl} = \|\mathbf{u}_k - \mathbf{u}_l\|$, and $\mathcal{S}_K$ denotes the symmetric group on $K$ elements. Eq.~\eqref{eq:blind_matching} is solved with the factorized Hahn-Grant relaxation of~\cite{schnaus2025s}. The optimum $\pi^\star$ assigns each visual cluster a category label without any paired vision-language data, contrastive pretraining, or VLM supervision; the alignment relies only on the shared relational geometry asserted by Eq.~\eqref{eq:platonic-alignment}.

\paragraph{Goal localization and semantic path planning.}
Given a query category $c_t$, the corresponding visual cluster index is $k^\star = (\pi^\star)^{-1}(t)$, and the candidate goal set is $S_{k^\star} = \{i : \sigma(i) = k^\star\}$. When multiple instances of the queried category coexist in the scene (e.g., several chairs), we treat $S_{k^\star}$ as a set of admissible goals, consistent with the open-ended interpretation of ObjNav success (``reach \emph{any} chair''). Goal selection and path planning are unified: starting from the agent's currently localized node $i_0$, we run Dijkstra over $\mathcal{G}$ under the hybrid edge weight $d(\cdot,\cdot)$ and return the path to the admissible node with minimum accumulated cost,
\begin{equation}
    g \;=\; \arg\min_{i \in S_{k^\star}} \mathrm{cost}_d(i_0, i),
\end{equation}
where $\mathrm{cost}_d(i_0, i)$ is the Dijkstra path cost from $i_0$ to $i$ under $d$. Compared with conventional planning that prioritizes shortest geometric paths, this formulation favors trajectories that traverse semantically meaningful object transitions while remaining grounded in the underlying geometry.

\paragraph{Real-world execution.}
For real-world deployment, we instantiate PTM under an RGB-only teach-and-repeat setting.
The complete segment-level map is constructed from all SAM2 segments, while DINOv3-based linear-probed background filtering is applied at execution time by removing edges incident to background nodes before Dijkstra planning.
The detailed robot setup is described in Section~\ref{sec:real_world}.

\section{Experiments}

\begin{table}[t]
\centering
\captionsetup{justification=centering,singlelinecheck=false}
\caption{Object goal navigation results on HM3D-IIN~\cite{krantz2022instance} and HM3D-OVON~\cite{yokoyama2024hm3dovon}. \\
  {\small \textbf{IIN:} PlatonicNav with goal grounded by GT mask achieves higher SPL and SSPL than ObjectReact. \textbf{OVON:} PlatonicNav outperforms the vast majority of cross-modal ObjNav methods on both SR and SPL.}}
\label{tab:hm3d-ovon}
\small
\setlength{\tabcolsep}{3.2pt}
\begin{tabular}{lcccccccc}
\toprule
 & \multicolumn{2}{c}{HM3D-IIN} 
 & \multicolumn{2}{c}{Val-Seen} 
 & \multicolumn{2}{c}{Val-Seen-Synonyms} 
 & \multicolumn{2}{c}{Val-Unseen} \\
\cmidrule(lr){2-3} \cmidrule(lr){4-5} \cmidrule(lr){6-7} \cmidrule(lr){8-9}
Method 
& SPL$\uparrow$ & SSPL$\uparrow$
& SR$\uparrow$ & SPL$\uparrow$
& SR$\uparrow$ & SPL$\uparrow$
& SR$\uparrow$ & SPL$\uparrow$ \\
\midrule
ObjectReact ~\cite{garg2025objectreact} & 59.1 & 64.6 & -- & -- & -- & -- & -- & -- \\
BC~\cite{yokoyama2024hm3dovon}         & -- & -- & 11.1 & 4.5  & 9.9  & 3.8  & 5.4  & 1.9  \\
DAgger~\cite{ross2011dagger}     & -- & -- & 11.1 & 4.5  & 9.9  & 3.8  & 5.4  & 1.9  \\
RL~\cite{schulman2017proximal}         & -- & -- & 18.1 & 9.4  & 15.0 & 7.4  & 10.2 & 4.7  \\
BCRL~\cite{schulman2017proximal}       & -- & -- & 39.2 & 18.7 & 27.8 & 11.7 & 18.6 & 7.5  \\
DAgRL~\cite{yokoyama2024hm3dovon}       & -- & -- & 41.3 & 21.2 & 29.4 & 14.4 & 18.3 & 7.9  \\
VLFM\textsuperscript~\cite{yokoyama2024vlfm}      & -- & -- & 35.2 & 18.6 & 32.4 & 17.3 & 35.2 & 19.6 \\
DAgRL+OD~\cite{yokoyama2024hm3dovon}   & -- & -- & 38.5 & 21.1 & 39.0 & 21.4 & 37.1 & 19.8 \\
Uni-NaVid\textsuperscript~\cite{zhang2024uni} & -- & -- & 41.3 & 21.1 & 43.9 & 21.8 & 39.5 & 19.8 \\
MTU3D\textsuperscript~\cite{zhu2025mtu3d}     & -- & -- & 55.0 & 23.6 & 45.0 & 14.7 & 40.8 & 12.1 \\
\midrule
\textbf{PlatonicNav (Ours)}
& \textbf{62.6} & \textbf{70.8}
& \textbf{51.8} & \textbf{23.8}
& \textbf{49.8} & \textbf{23.1}
& \textbf{48.7} & \textbf{22.0} \\
\bottomrule
\end{tabular}
\end{table}

\begin{wraptable}{r}{0.5\textwidth}
\centering
\captionsetup{justification=raggedright,singlelinecheck=false,font=small}
\caption{Navigation results on R2R-CE Val-Unseen. \\
  {\small PlatonicNav outperforms a considerable portion of VLN baselines while there are still some VLN methods hold the lead on this benchmark.}}
\label{tab:r2rce-vln-baselines}

\begingroup
\scriptsize
\setlength{\tabcolsep}{2.0pt}
\renewcommand{\arraystretch}{0.88}

\resizebox{0.5\textwidth}{!}{%
\begin{tabular}{lcc@{\hspace{0.8em}}lcc}
\toprule
\multicolumn{3}{c}{R2R-CE Val-Unseen} &
\multicolumn{3}{c}{R2R-CE Val-Unseen} \\
\cmidrule(lr){1-3}\cmidrule(lr){4-6}
Method & SR$\uparrow$ & SPL$\uparrow$ &
Method & SR$\uparrow$ & SPL$\uparrow$ \\
\midrule
CMA~\cite{krantz2020beyond}        & 41.0 & 36.0 &
CM2~\cite{georgakis2022cm2}        & 34.0 & 27.0 \\
Sim2Sim~\cite{krantz2022sim2sim}   & 43.0 & 36.0 &
WS-MGMap~\cite{chen2022wsmgmap}    & 38.0 & 34.0 \\
GridMM~\cite{wang2023gridmm}       & 49.0 & 41.0 &
AO-Planner~\cite{chen2025aoplanner} & 47.0 & 33.0 \\
Ego$^2$-Map~\cite{hong2023ego2map} & 47.0 & 41.0 &
Seq2Seq~\cite{krantz2020beyond}    & 25.0 & 22.0 \\
DreamWalker~\cite{wang2023dreamwalker} & 49.0 & 44.0 &
CMA~\cite{krantz2020beyond}        & 32.0 & 30.0 \\
Reborn~\cite{an2022rxrhabitat}     & 50.0 & 46.0 &
NaVid~\cite{zhang2024navid}        & 37.0 & 35.0 \\
ETPNav~\cite{an2024etpnav}         & 57.0 & 49.0 &
Uni-NaVid~\cite{zhang2025uninavid} & 47.0 & 42.7 \\
HNR~\cite{wang2024hnr}             & 61.0 & 51.0 &
NaVILA~\cite{cheng2024navila}      & 54.0 & 49.0 \\
AG-CMTP~\cite{chen2021cmtp}        & 23.0 & 19.0 &
VLN-R1~\cite{qi2025vlnr1}          & 30.2 & 21.8 \\
R2R-CMTP~\cite{chen2021cmtp}       & 26.0 & 22.0 &
OctoNav~\cite{gao2025octonav}      & 37.1 & 33.6 \\
InstructNav~\cite{long2024instructnav} & 31.0 & 24.0 &
StreamVLN~\cite{wei2025streamvln}  & 56.9 & 51.9 \\
LAW~\cite{raychaudhuri2021law}     & 35.0 & 31.0 &
CorrectNav~\cite{yu2025correctnav} & 65.1 & 62.3 \\
\midrule
\textbf{PlatonicNav (Ours)} &  &  &
 & \textbf{63.5} & \textbf{40.4} \\
\bottomrule
\end{tabular}
}

\endgroup
\vspace{-1.0em}
\end{wraptable}

\label{sec:experiments}
\subsection{Benchmarks and Metrics}

\paragraph{Benchmarks.}
To thoroughly evaluate PlatonicNav, we adopt three complementary embodied navigation benchmarks. For comparing with vision-only Object Goal Navigation, we evaluate on \emph{HM3D-IIN}~\cite{krantz2022instance}, where the agent navigates to a target object instance in photorealistic \emph{HM3D}~\cite{ramakrishnan2021hm3d} scenes. For comparing with cross-modal-training Object Goal Navigation baselines, we adopt \emph{HM3D-OVON}~\cite{yokoyama2024hm3dovon}, where the goal is specified by an object category under an open-vocabulary setting. For further evaluating the generalization across independent modalities without explicit vision-language supervision, we  also adopt \emph{R2R-CE}~\cite{krantz2020beyond}, the continuous-environment version of Room-to-Room navigation, where the agent follows natural-language route instructions in unseen \emph{MP3D}~\cite{Matterport3D} environments.

\paragraph{Metrics.}
For \emph{IIN}, we report success weighted by path length (\emph{SPL}~\cite{anderson2018evaluation}) and soft success weighted by path length (\emph{SSPL})~\cite{batra2020objectnav}, following prior instance navigation evaluation protocols. \emph{SPL} measures whether the agent successfully reaches the goal while penalizing inefficient trajectories, whereas \emph{SSPL} further accounts for partial progress toward the target when the episode is not strictly successful. For \emph{OVON} and \emph{R2R-CE}, we report success rate (\emph{SR}) and \emph{SPL}. \emph{SR} measures the percentage of episodes in which the agent stops within the task-specific success threshold, while \emph{SPL} jointly measures goal-reaching success and path efficiency.

\paragraph{Implementation details.}
For \emph{IIN}, we follow the same purely visual goal-selection protocol as ObjectReact~\cite{garg2025objectreact} to minimize the influence of cross-modal grounding errors and isolate the evaluation of the Platonic Topological Map and downstream PlatonicObject Costmap with $(\lambda_g, \lambda_s)= (0.8, 0.2)$. For \emph{OVON} and \emph{R2R-CE}, we chose short-horizon subsets, implementing full PlatonicNav pipeline including mapping, blind-match goal selection, and execution as illustrated in Fig~\ref{fig:top-pipeline}.  

\paragraph{Main results.} 
\textbf{IIN: } On HM3D-IIN, PlatonicNav achieves higher SPL and SSPL than ObjectReact, reflecting the contribution of the Platonic Topological Map rather than differences in cross-modal grounding. In particular, the comparison suggests that augmenting purely geometric edge weights with semantic distances from a self-supervised visual encoder enables the agent to reason over object relations in a more semantically structured manner, leading to more efficient navigation behavior. This also aligns with the observation of their trajectory comparison (see Appendix~\ref{sec: top-down-tra-com}) 
\textbf{OVON: } On HM3D-OVON, PlatonicNav outperforms multiple cross-modal-training ObjNav. This result indicates that explicit vision-language supervision is not the only way to ground language-specified object goals in embodied navigation. Instead, the relational structure captured by independently trained vision and language encoders can be exploited through blind matching, supporting that the two modalities share an implicit semantic structure that is useful for navigation. 
\textbf{R2R-CE: } On R2R-CE, PlatonicNav still outperforms many VLN baselines. Together with the OVON results, this further suggests that explicit cross-modal training is not a necessary condition for connecting language goals with visual navigation representations. 
Taken together, the three experiments provide support for our central proposition: the \emph{Platonic Representation Hypothesis} (Sec~\ref{sec:platonic_topomap} ) can be operationalized in embodied navigation. Specifically, PlatonicNav uses the \emph{Platonic Topological Map} as a shared interface to connect vision-only ObjNav, cross-modal ObjNav, and VLN under a unified semantic navigation framework.

\paragraph{Qualitative simulation results.}
We provide additional qualitative simulation visualizations in Appendix~\ref{app:ablation_results}.
The appendix first presents VLN simulation examples in Figs.~\ref{fig:ablation_vln_bottom_stairs}--\ref{fig:ablation_vln_office_desk}, followed by ObjNav simulation examples in Figs.~\ref{fig:ablation_platonic_refrigerator}--\ref{fig:ablation_platonic_photo}.
These examples show temporally ordered ego-view, depth, and BEV observations across different navigation tasks.

\subsection{Real-World Evaluation}
\label{sec:real_world}

\paragraph{Unitree Go2 Platform.}
Our second platform is the \textbf{Unitree Go2} quadruped robot, a more advanced and robust system designed for real-world deployment, providing strong locomotion stability and rich geometric sensing.

\paragraph{Go2 Air deployment details.}
We deploy PlatonicNav on a Unitree Go2 Air quadruped with an onboard RGB camera.
The high-level pipeline runs off-board on a laptop via a ROS2 interface based on \texttt{go2\_ros2\_sdk}.
The laptop runs Ubuntu 22.04.5 LTS with an NVIDIA GeForce RTX 4060 Laptop GPU.
RGB observations are streamed from \texttt{/camera/image\_raw}, and the online worker publishes \texttt{geometry\_msgs/Twist} commands to \texttt{/cmd\_vel\_out}.

The real-world map is constructed offline from a human-guided teaching trajectory and reused during repeat execution.
We use a teach-and-repeat goal specification by setting the goal to the terminal topological node of the teaching trajectory, i.e., \texttt{goalNodeIdx=-1}.
This avoids simulator-only ground-truth masks and focuses the evaluation on PTM-based real-world planning and control.
The ObjectReact controller performs RGB-only control before sending commands to the Go2 native locomotion controller.

\paragraph{Qualitative real-world evaluation.}
We provide qualitative visualizations for both ObjectNav and VLN real-world executions.
For ObjectNav, Figs.~\ref{fig:objnav_task1_teach}, \ref{fig:objnav_task1_repeat}, \ref{fig:objnav_task2_teach}, \ref{fig:objnav_task2_repeat}, \ref{fig:objnav_task3_teach}, and \ref{fig:objnav_task3_repeat} show three teach-and-repeat tasks.
For VLN, Fig.~\ref{fig:vln_teach} shows the teach phase, while Figs.~\ref{fig:vln_repeat_lamp}, \ref{fig:vln_repeat_plant}, and \ref{fig:vln_repeat_chair} show repeat executions under the instructions \textit{go to the lamp}, \textit{find the plant}, and \textit{go to the chair}.
Each visualization presents temporally ordered ego-view, estimated depth, and pose-aligned point-map observations.

\section{Conclusion}

We propose \emph{PlatonicNav}, a representation-centric framework for embodied navigation that unifies not only cross-modal-training Object Goal Navigation and vision-only Object Goal Navigation, but also Vision-Language Navigation through the \emph{Platonic Representation Hypothesis}. Instead of relying on explicit visual-language supervision or architectural unification, our approach leverages the intrinsic semantic alignment between independently trained visual and language models. 
Building upon this insight, we introduce \emph{Platonic Topological Maps}, where navigation is formulated as geodesic traversal over a learned semantic manifold rather than purely geometric space. This perspective reinterprets topological maps as discretizations of representation space, enabling both object-driven and language-driven navigation within a single unified framework.
Extensive experiments across representative benchmarks and real-world robot platforms demonstrate that our method generalizes across tasks, modalities, and sensing configurations. These results suggest that semantic structure in representation space provides a principled foundation for embodied navigation, opening new directions toward unified, scalable, and modality-agnostic navigation systems.

\clearpage
\bibliographystyle{plain} 
\bibliography{references}

\clearpage
\appendix

\section{Preliminaries}

\subsection{Platonic Representation Hypothesis}
\label{sec: prelim-platonic}
\begin{formal}
\textbf{The Platonic Representation Hypothesis~\cite{huh2024platonic}.}
Neural networks, trained with different objectives
on different data and modalities, are converging to a
shared statistical model of reality in their representation spaces.
\end{formal}

We adopt the formulation of Huh et al.~\cite{huh2024platonic}. Let $f_v: \mathcal{I} \to \mathbb{R}^{d_v}$ be a visual encoder trained with self-supervised objectives (e.g., DINOv3~\cite{simeoni2025dinov3}), and let $f_l: \mathcal{T} \to \mathbb{R}^{d_l}$ be a language encoder trained on large-scale text corpora (e.g., the GTR-T5 dense retriever~\cite{ni-etal-2022-large}). Given $N$ concepts realized as image samples $\{x_i\}_{i=1}^{N}$ on the visual side and textual descriptions $\{c_i\}_{i=1}^{N}$ on the language side, define the pairwise distance matrices
\begin{equation}
    D^{v}_{ij} = d_v\!\big(f_v(x_i),\, f_v(x_j)\big),
    \qquad
    D^{l}_{ij} = d_l\!\big(f_l(c_i),\, f_l(c_j)\big),
\end{equation}
where $d_v(\cdot,\cdot)$ and $d_l(\cdot,\cdot)$ are distances in the respective embedding spaces. The hypothesis asserts that, after a normalization $\mathcal{N}(\cdot)$ that removes modality-specific scale,
\begin{equation}
    \mathcal{N}(D^{v}) \;\approx\; \mathcal{N}(D^{l}),
    \label{eq:platonic-alignment}
\end{equation}
even though $f_v$ and $f_l$ are trained without any cross-modal supervision, such as CLIP-style contrastive pretraining~\cite{radford2021learning}, vision-language models~\cite{bai2025qwen3}, or paired image-text data. Intuitively, the relative geometry of concepts is preserved across modalities (Fig.~\ref{fig:blind-match}).

Motivated by Eq.~\eqref{eq:platonic-alignment}, \emph{blind matching}~\cite{schnaus2025s} recovers cross-modal correspondences between a set of visual embeddings and a set of language embeddings by aligning their pairwise distance matrices, without any paired data or contrastive training. We use this property to ground language goals into a vision-only topological map (Section~\ref{sec:platonic_topomap}).

\subsection{Topological Map for Embodied Navigation}
\label{sec:topo-map-for-em}
\begin{wrapfigure}{r}{0.5\linewidth}
    \centering
    \vspace{-5pt}
    \includegraphics[width=\linewidth]{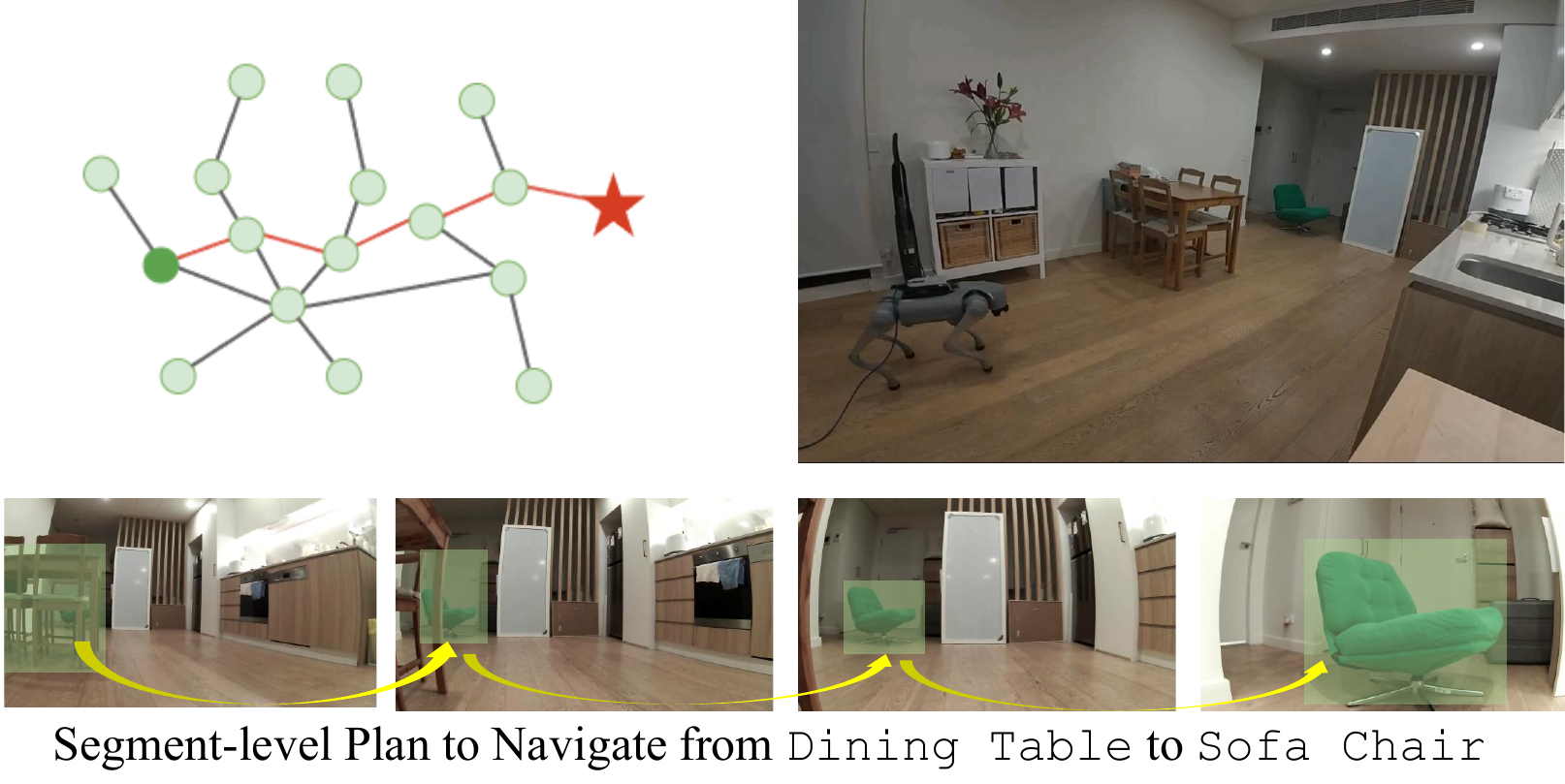}
    \vspace{-10pt}
    \caption{
    \textbf{Segment-based topological map.}
    Image segments serve as graph nodes, and navigation is planned as a sequence of segment-level ``hops'' over a sparse graph. Figure adapted from~\cite{garg2024robohop}.
    }
    \vspace{-10pt}
    \label{fig:segment-based}
\end{wrapfigure}

A topological map represents an environment as a graph $\mathcal{G} = (\mathcal{V}, \mathcal{E})$, where each node $v \in \mathcal{V}$ corresponds to an observation, landmark, object, or spatial region, and each edge $e \in \mathcal{E}$ encodes connectivity or traversal cost. Compared with dense metric mapping such as SLAM-based point-cloud reconstruction, this graph-structured abstraction trades global geometric consistency for sparsity and structure-awareness. RoboHop~\cite{garg2024robohop} instantiates the abstraction as a \emph{segment-based topological map}, in which nodes are image segments and edges encode spatial relations among them, enabling planning as a sequence of segment-level hops (Fig.~\ref{fig:segment-based}).

ObjectReact~\cite{garg2025objectreact} extends this paradigm into an \emph{object-relative navigation pipeline}. The environment is modeled as a topometric graph whose nodes are object-centric image segments, intra-image edges encode relative 3D geometric distances, and inter-image edges are formed by cross-view object association. At inference, given a goal segmentation mask, the agent localizes the target by selecting the map segment with the highest mask intersection-over-union, computes shortest-path distances over $\mathcal{G}$ to the goal node, and projects these distances back onto the input image as a dense \emph{WayObject Costmap} that is consumed by a downstream control policy.

\section{Meter-scale calibration of Platonic distance}
\label{sec:d_plat_meter}
The Platonic distance $d_{\mathrm{plat}}(i,j)$ in Eq.~\eqref{eq:d_plat} is a
cosine distance between visual embeddings and is therefore dimensionless, while
$d_{\mathrm{geo}}(i,j)$ is ObjectReact's geometric edge cost measured in meters.
Directly mixing these two quantities would change the scale of graph
shortest-path distances received by the frozen ObjectReact controller. We
therefore calibrate $d_{\mathrm{plat}}$ to the same meter scale as
$d_{\mathrm{geo}}$ before computing the hybrid edge weight in
Eq.~\eqref{eq:hybrid_edge_weight}.

For each episode graph, let $\mathcal{E}_{\mathrm{nav}}$ denote the eligible
navigation edges used for scale estimation. We exclude ObjectReact
inter-image and same-object association edges from this set, since these edges
encode object association rather than local navigation cost and keep their
original geometric cost. We estimate robust per-episode scales by the 95th
percentile:
\begin{equation}
s_{\mathrm{geo}}
=
Q_{95}\!\left(
\left\{
d_{\mathrm{geo}}(i,j)
\mid
(i,j)\in\mathcal{E}_{\mathrm{nav}}
\right\}
\right).
\end{equation}
\begin{equation}
s_{\mathrm{plat}}
=
Q_{95}\!\left(
\left\{
d_{\mathrm{plat}}(i,j)
\mid
(i,j)\in\mathcal{E}_{\mathrm{nav}}
\right\}
\right).
\end{equation}

We then convert the dimensionless Platonic distance into a meter-scale penalty:
\begin{equation}
\label{eq:d_plat_meter}
\tilde{d}_{\mathrm{plat}}(i,j)
=
s_{\mathrm{geo}}
\cdot
\mathrm{clip}
\left(
\frac{d_{\mathrm{plat}}(i,j)}{s_{\mathrm{plat}}},
0,
c
\right),
\end{equation}
where
\begin{equation}
\mathrm{clip}(x,0,c)=\min(\max(x,0),c).
\end{equation}
In all reported experiments we use $c=2.0$, which limits rare embedding-distance
outliers from dominating the graph cost. If a scale estimate is non-finite or
degenerate, we fall back to a positive default scale and record this in the graph
metadata.

With this calibration, $\tilde{d}_{\mathrm{plat}}(i,j)$ has units of meters, so
the final hybrid edge weight
\begin{equation}
d(i,j)
=
\lambda_g d_{\mathrm{geo}}(i,j)
+
\lambda_s \tilde{d}_{\mathrm{plat}}(i,j)
\end{equation}
remains on the same order-of-magnitude meter scale as ObjectReact's original
edge weight. Setting $\lambda_g=1$ exactly recovers the original ObjectReact
geometric graph.

\section{Top-down Trajectory Comparison}
\label{sec: top-down-tra-com}
\begin{center}
    \includegraphics[width=\linewidth]{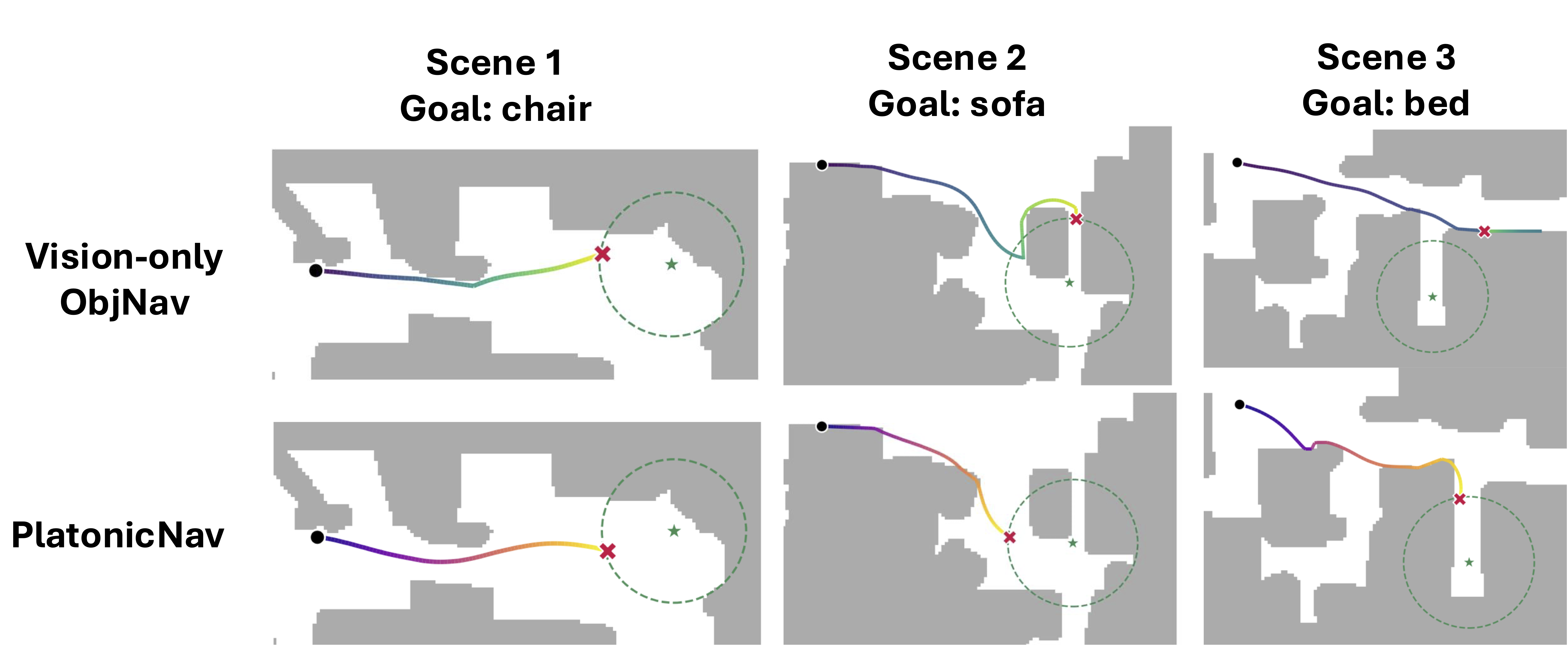}
    \captionof{figure}{
    \textbf{Top-down trajectory map of vision-only ObjNav and PlatonicNav  on \emph{HM3D-IIN}.}
    We visualize the navigation trajectories of vision-only ObjNav (e.g., ObjectReact~\cite{garg2025objectreact}) and PlatonicNav with pure vision goal grounding. Their trajectories shows relative similarity while PlatonicNav's trajectories seem more straightforward than ObjectReact's.
    }
    \label{fig:Pla-Obj-tra-comp}
\end{center}
Observing both similarity and difference between vision-only ObjNav's trajectories and PlatonicNav's trajectories, we find that even though their shapes are generally similar, PlatonicNav acts more efficiently than vision-only ObjNav in most scenes. This difference highly align with the result of main experiment on \emph{IIN}: by injecting language-level semantic information to pure vision modality, \emph{Platonic Topological Map} makes navigation agent's actions more semantic-meaningful.

\section{Real-world Implementation Platform}
\begin{center}
    \includegraphics[width=\linewidth]{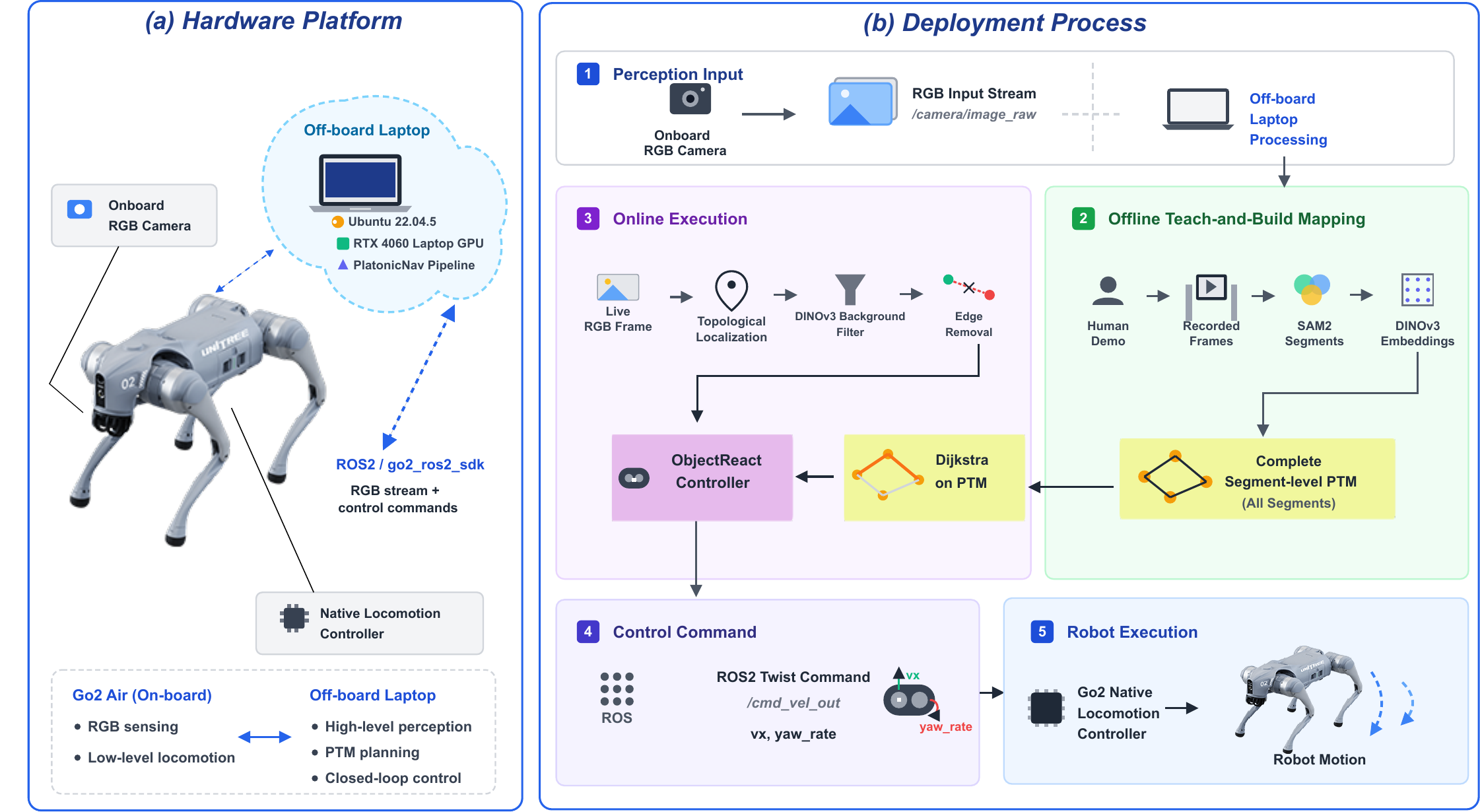}
    \captionof{figure}{
    \textbf{Real-world robot platforms for evaluation.}
    We deploy our method on a quadruped Unitree Go2 robot, providing robust perception and locomotion. These platforms demonstrate the applicability of Platonic Topological Maps in embodied system.
    }
    \label{fig:real_robot}
\end{center}

\paragraph{Evaluation Protocol.}
For both platforms, we construct topological maps from onboard sensory inputs and evaluate navigation performance under object-goal and language-conditioned scenarios. 
The experiments are designed to test whether semantic distances in representation space can effectively guide real-world navigation, even under noisy observations and limited sensing conditions.

\section{Limitation and future work}
\label{sec: limitation}
\paragraph{Limitation. } PlatonicNav is intended as an initial step toward validating the Platonic Representation Hypothesis in Embodied Navigation, rather than a fully optimized end-to-end navigation system. Its current performance is bounded by several modular components, including the quality of visual segmentation, the robustness of blind matching, the expressiveness of the language encoder, and the long-distance capability of the \emph{ObjectReact}-style controller~\cite{garg2025objectreact}. In particular, the R2R-CE results suggest that handling long natural-language instructions remains challenging.

\paragraph{Future work. } Future work will focus on improving the robustness of vision-language matching, strengthening goal extraction from complex instructions, and post-training the controller for long-distance navigation. We also plan to further optimize the overall architecture to better support open-vocabulary, long-context embodied navigation.

\section{Ablation Study}

\subsection{Selection of metric for \emph{Platonic distance}}
\label{sec:raw_d_plat_ablation}
To determine which metric is more suitable for \emph{Platonic distance}, we compare the SPL and SSPL of \emph{Platonic Topoligical Map} with different metrics for \emph{Platonic distance} under the same configuration on \emph{HM3D-IIN}. Table~\ref{tab:d_plat_metrics} shows that using L2-normalized cosine distance(shown as Eq.~\eqref{eq:d_plat}) as the metric of \emph{Platonic distance} outperforms L2-normalized euclidean distance, indicating that cosine distance has the better ability of capturing semantic information from visual embedding space. Also, better performance of PTM with L2-normalized visual embeddings suggests that L2-normalization removes feature-norm effects, so the Platonic distance reflects semantic direction similarity rather than raw embedding magnitude. Taking these two aspects into account, L2-normalized cosine distance is chosen to compute the \emph{Platonic distance}.

\begin{table}[t]
\centering
\caption{Comparison of PTM with different \emph{Platonic distance} metrics on \emph{HM3D-IIN}}
\begin{tabular}{ccc}
\toprule
    Metric & SPL & SSPL \\
\midrule
    L2-normalized Cosine Distance & 62.6 & 70.8 \\
    L2-normalized Euclidean Distance & 58.4 & 69.7 \\
    Raw Euclidean Distance & 57.5 & 68.3 \\
\bottomrule
\end{tabular}
\label{tab:d_plat_metrics}
\vspace{-1em}
\end{table}

\begin{table}[t]
\centering
\small

\begin{minipage}{0.48\linewidth}
\centering
\caption{Comparison of PTM with different $(\lambda_g, \lambda_s)$ pairs on \emph{HM3D-IIN}}
\begin{tabular}{cccc}
\toprule
$\lambda_g$ & $\lambda_s$ & SPL & SSPL \\
\midrule
1.00 & 0.00 & 59.1 & 64.6 \\
0.95 & 0.05 & 59.7 & 65.5 \\
0.90 & 0.10 & 60.1 & 67.5 \\
0.80 & 0.20 & 62.6 & 70.8 \\
0.50 & 0.50 & 56.3 & 62.9 \\
\bottomrule
\end{tabular}
\label{tab:lambda_ablation}
\end{minipage}
\hfill
\begin{minipage}{0.48\linewidth}
\centering
\caption{Comparison of Segmentation on \emph{HM3D-IIN}}
\begin{tabular}{cccc}
\toprule
Method & Segmentation & SPL & SSPL \\
\midrule
ObjectReact & GT segmentation & 59.1 & 64.6 \\
ObjectReact & FastSAM~\cite{zhao2023fastsam} & 28.6 & 37.5 \\
PlatonicNav & GT segmentation & 62.6 & 70.8 \\
PlatonicNav & FastSAM~\cite{zhao2023fastsam} & 29.5 & 41.7 \\
PlatonicNav & SAM2~\cite{kirillov2023sam} & 39.7 & 46.6 \\
\bottomrule
\end{tabular}
\label{tab:segmentation_ablation}
\end{minipage}
\vspace{-1em}
\end{table}

\subsection{Selection of $(\lambda_g, \lambda_s)$}
\label{sec:lambda_ablation}
We compare the Platonic Topological Map with the edge weights calculated from different $(\lambda_g, \lambda_s)$ pairs, evaluating them on \emph{HM3D-IIN}. From the Tab~\ref{tab:lambda_ablation}, we can find that the Platonic Topological Map achieve the best performance under $(\lambda_g, \lambda_s) = (0.8, 0.2)$ configuration, indicating that injecting semantic information to pure geometric edge weight does improve the efficiency of navigation. At the mean time, we can also notice that, when Platonic Topological Map is with $(\lambda_g, \lambda_s) = (0.8, 0.2)$, both two metrics are lower than pure geometric topological map, suggesting that excess semantic weight might be detrimental to navigation behavior.

\subsection{Comparison of Segmentation}
To further explore the effect of segmentation quality in our navigation task, we compare ObjectReact and PlatonicNav under different segmentation configurations. As shown in Tab~\ref{tab:segmentation_ablation}, with the ground truth segmentation provided by \emph{IIN}, both ObjectReact and PlatonicNav achieve a relative high performance. However, when it comes to the segmentation provided by FastSAM~\cite{zhao2023fastsam}, the SPL and SSPL of both methods drop severely to nearly half. Mean while, we also notice that when PlatonicNav uses the segmentation from SAM2, its SPL increases by ten percentage points, and SSPL increases from $41.1$ to $46.6$. This results indicate that the quality of segmentation has a relatively strong impact on the navigation quality. In addition, the advantage of PlatonicNav with both ground truth segmentation and FastSAM segmentation further suggesting that the improvement of Platonic Topological Map comes from the injection of semantic information.

\section{Additional Real-world Qualitative Results}

\subsection{ObjectNav Qualitative Results}

\begin{center}
    \includegraphics[width=0.98\linewidth]{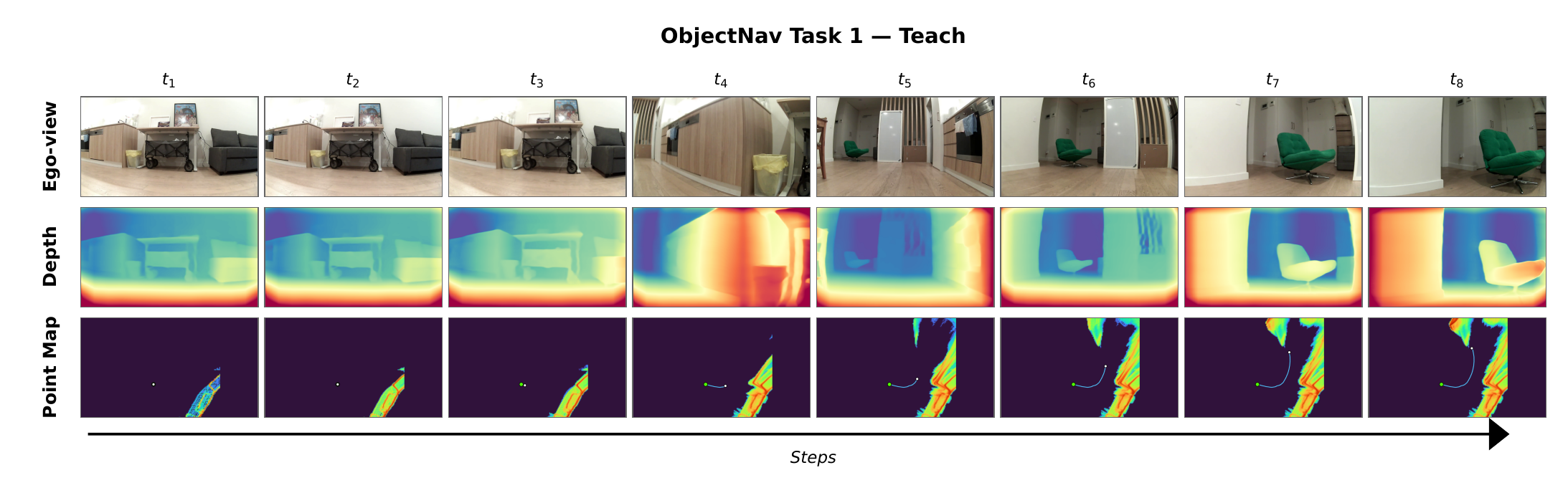}
    \captionof{figure}{\textbf{ObjectNav Task 1, teach phase.} Qualitative visualization.}
    \label{fig:objnav_task1_teach}
\end{center}

\begin{center}
    \includegraphics[width=0.98\linewidth]{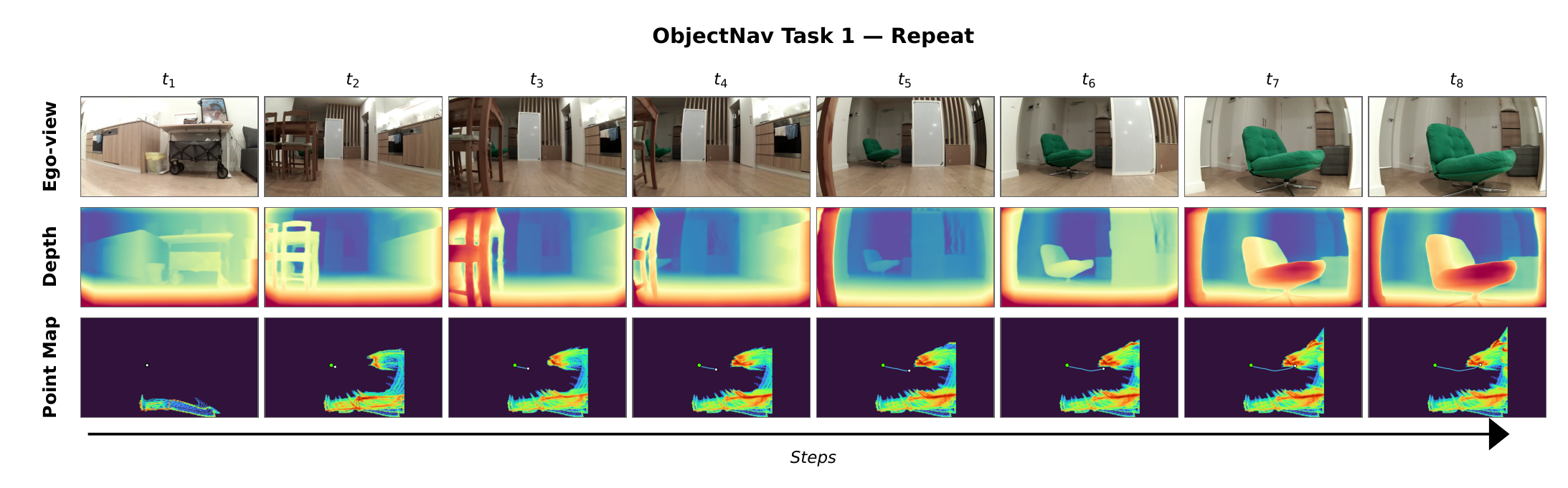}
    \captionof{figure}{\textbf{ObjectNav Task 1, repeat phase.} Qualitative visualization.}
    \label{fig:objnav_task1_repeat}
\end{center}

\begin{center}
    \includegraphics[width=0.98\linewidth]{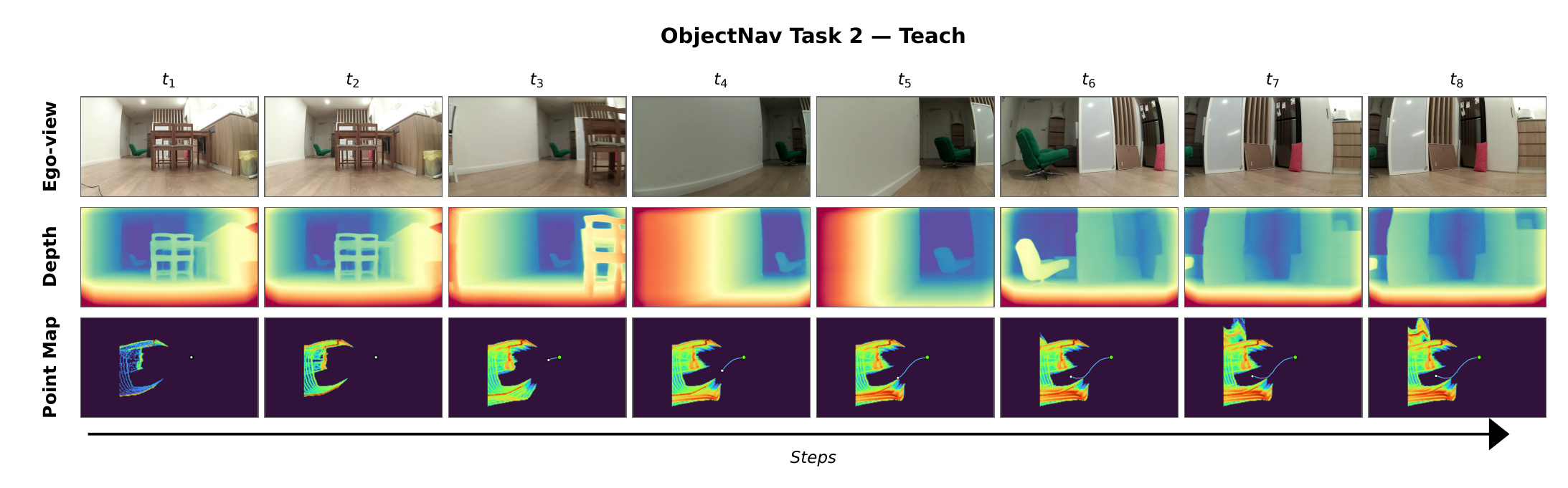}
    \captionof{figure}{\textbf{ObjectNav Task 2, teach phase.} Qualitative visualization.}
    \label{fig:objnav_task2_teach}
\end{center}

\begin{center}
    \includegraphics[width=0.98\linewidth]{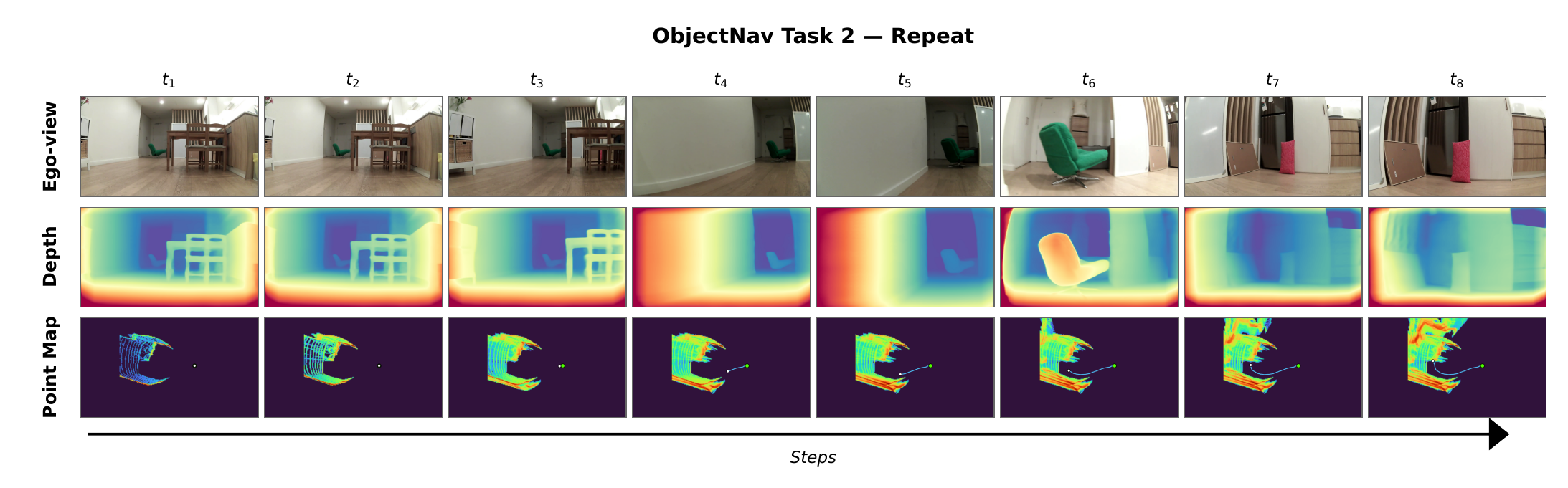}
    \captionof{figure}{\textbf{ObjectNav Task 2, repeat phase.} Qualitative visualization.}
    \label{fig:objnav_task2_repeat}
\end{center}

\begin{center}
    \includegraphics[width=0.98\linewidth]{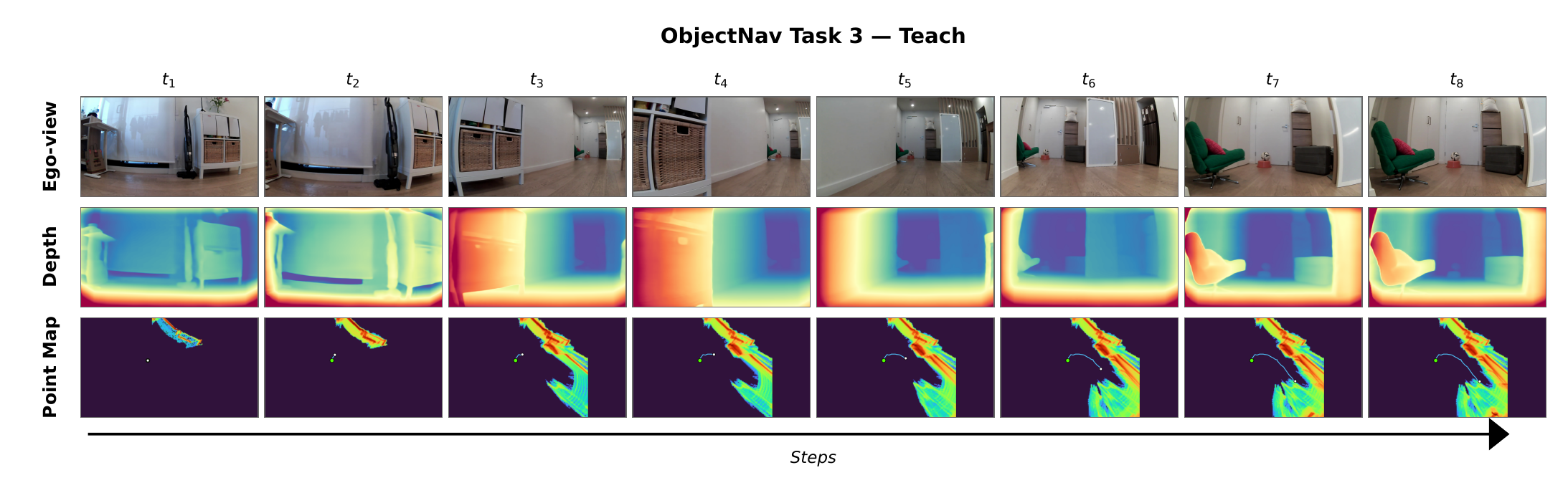}
    \captionof{figure}{\textbf{ObjectNav Task 3, teach phase.} Qualitative visualization.}
    \label{fig:objnav_task3_teach}
\end{center}

\begin{center}
    \includegraphics[width=0.98\linewidth]{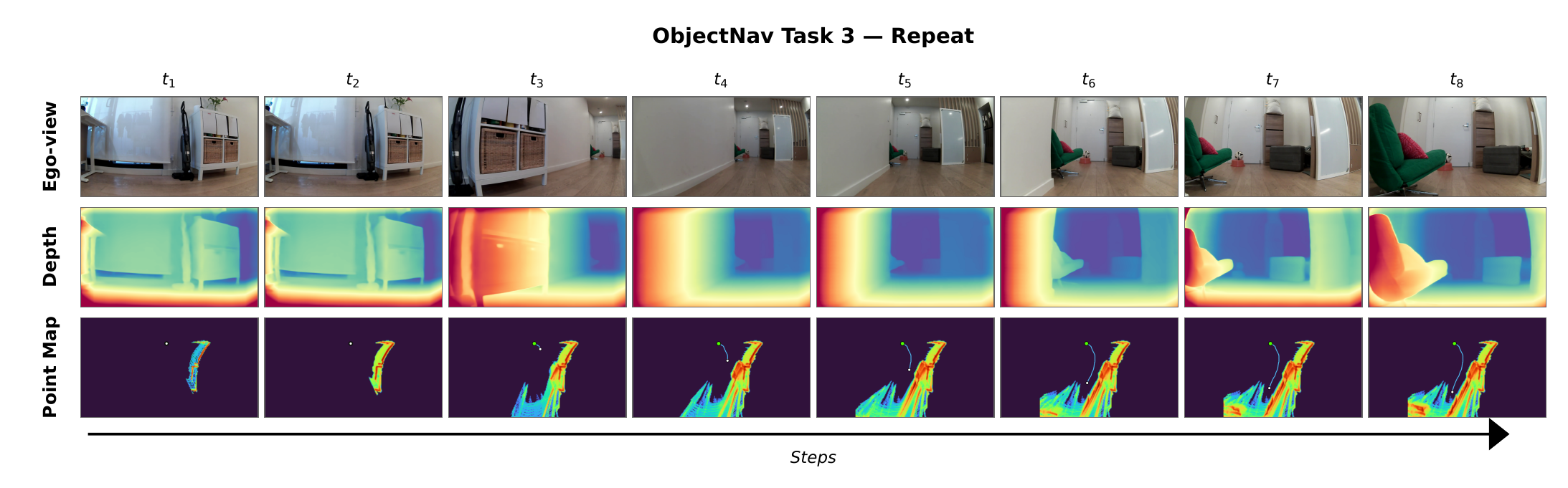}
    \captionof{figure}{\textbf{ObjectNav Task 3, repeat phase.} Qualitative visualization.}
    \label{fig:objnav_task3_repeat}
\end{center}

\subsection{VLN Qualitative Results}

\begin{center}
    \includegraphics[width=0.98\linewidth]{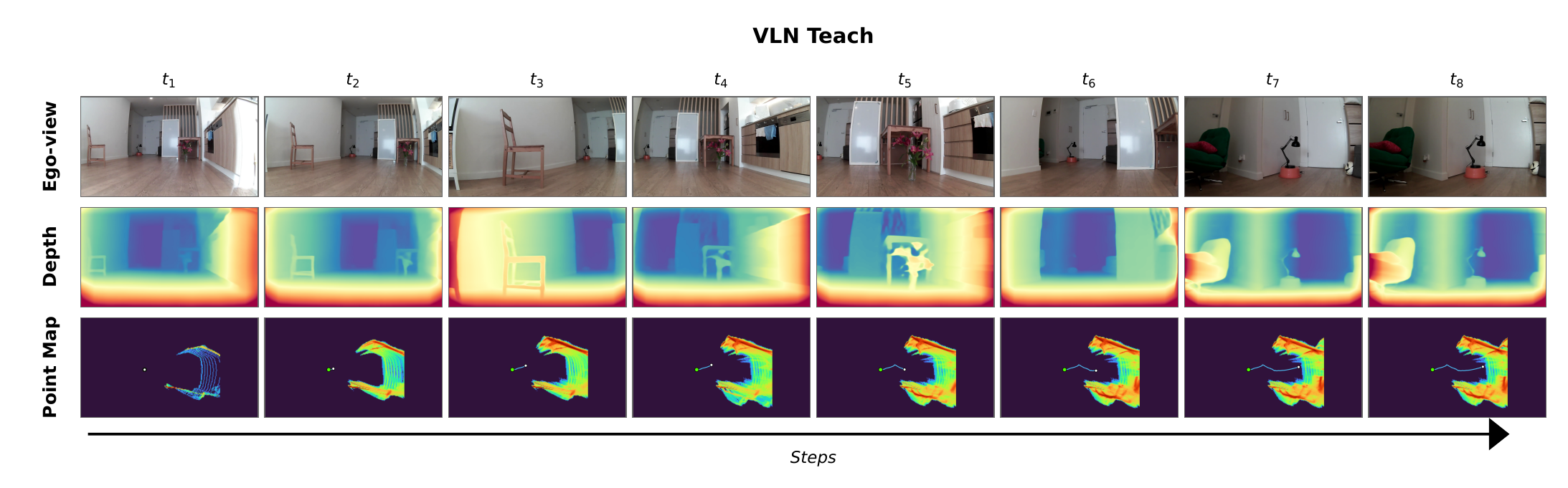}
    \captionof{figure}{\textbf{VLN teach phase.} Qualitative visualization.}
    \label{fig:vln_teach}
\end{center}

\begin{center}
    \includegraphics[width=0.98\linewidth]{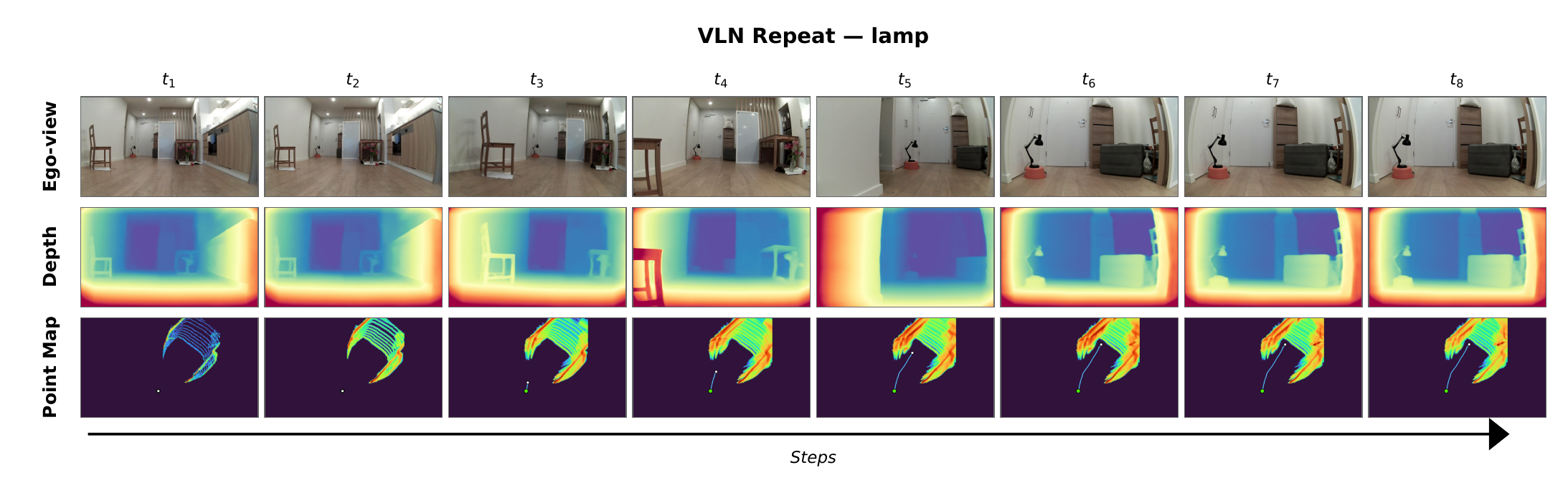}
    \captionof{figure}{\textbf{VLN repeat phase, go to the lamp.} Qualitative visualization.}
    \label{fig:vln_repeat_lamp}
\end{center}

\begin{center}
    \includegraphics[width=0.98\linewidth]{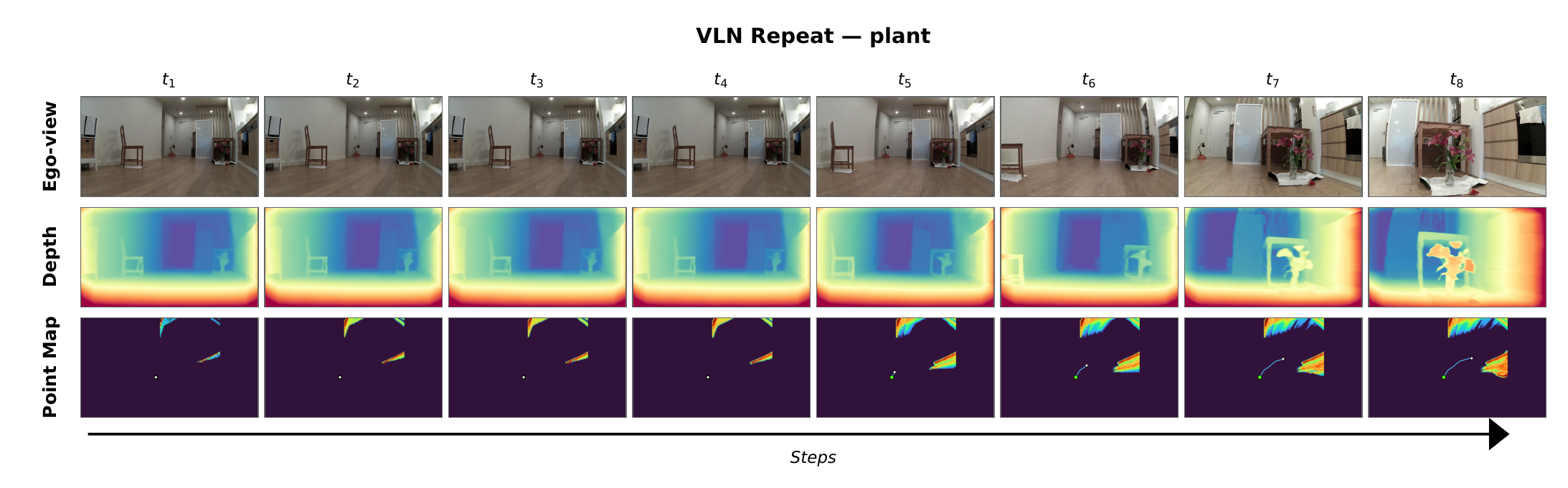}
    \captionof{figure}{\textbf{VLN repeat phase, find the plant.} Qualitative visualization.}
    \label{fig:vln_repeat_plant}
\end{center}

\begin{center}
    \includegraphics[width=0.98\linewidth]{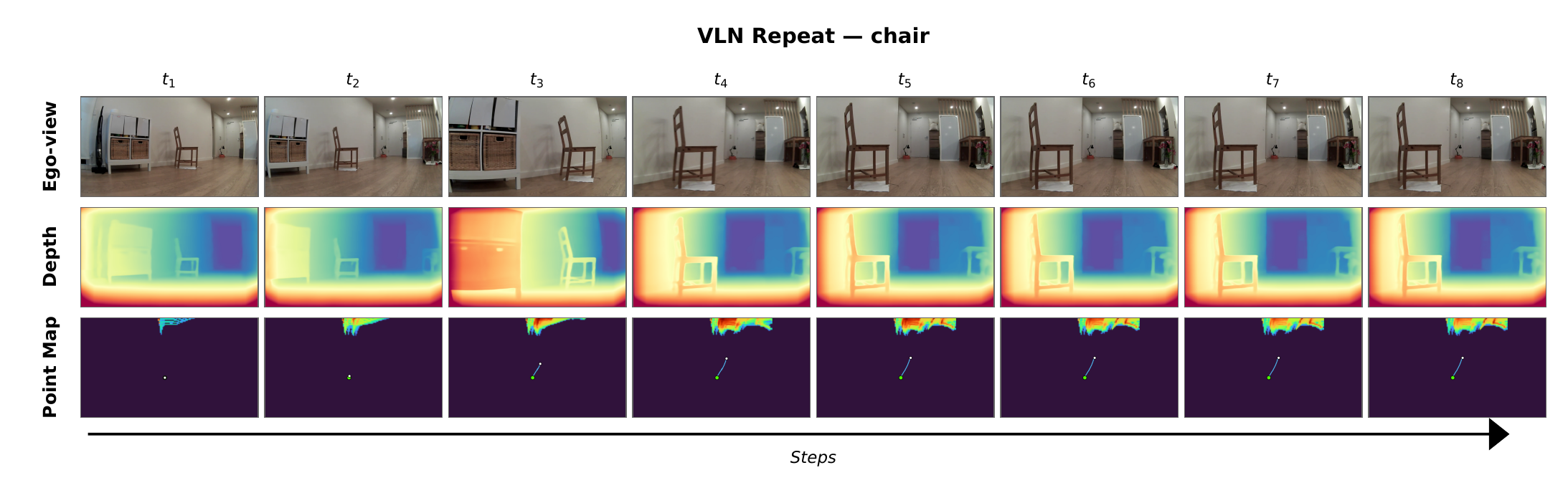}
    \captionof{figure}{\textbf{VLN repeat phase, go to the chair.} Qualitative visualization.}
    \label{fig:vln_repeat_chair}
    \vspace{-1em}
\end{center}

\section{Additional Simulation Results}
\label{app:ablation_results}

\subsection{VLN Simulation Results}

\begin{center}
    \begin{minipage}{0.98\linewidth}
        \centering
        \includegraphics[width=\linewidth]{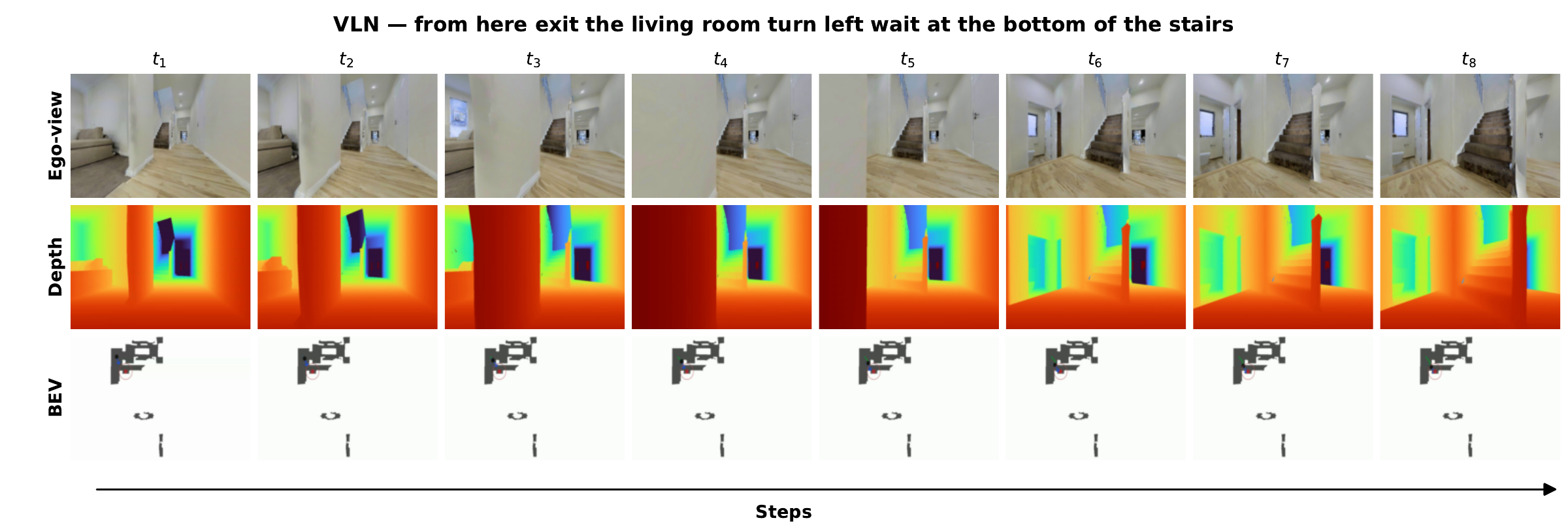}
        \captionof{figure}{\textbf{VLN simulation task, bottom of stairs.} Qualitative visualization.}
        \label{fig:ablation_vln_bottom_stairs}
    \end{minipage}
\end{center}

\begin{center}
    \begin{minipage}{0.98\linewidth}
        \centering
        \includegraphics[width=\linewidth]{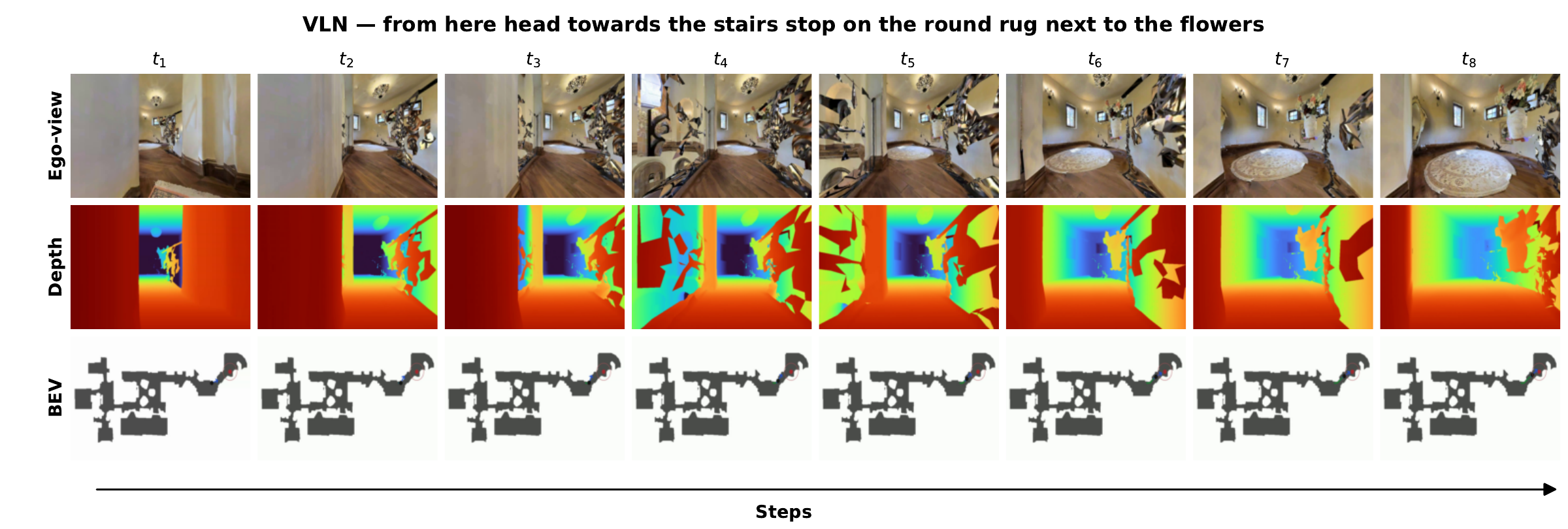}
        \captionof{figure}{\textbf{VLN simulation task, round rug near flowers.} Qualitative visualization.}
        \label{fig:ablation_vln_round_rug_flowers}
    \end{minipage}
\end{center}

\begin{center}
    \begin{minipage}{0.98\linewidth}
        \centering
        \includegraphics[width=\linewidth]{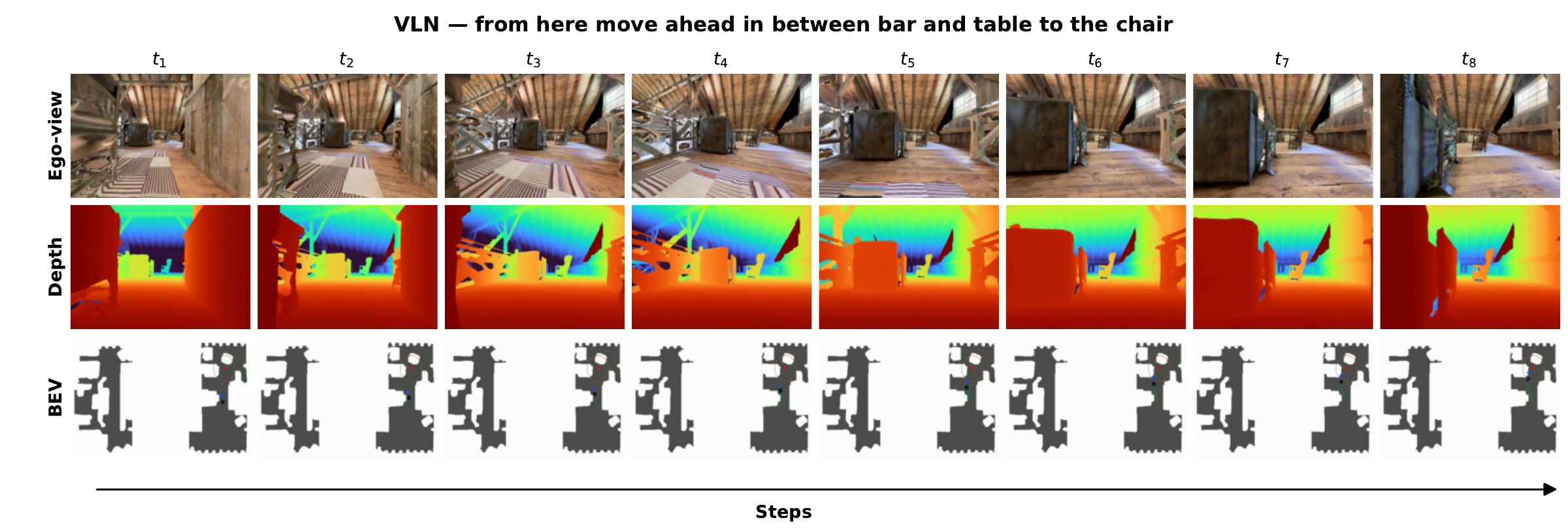}
        \captionof{figure}{\textbf{VLN simulation task, chair near bar and table.} Qualitative visualization.}
        \label{fig:ablation_vln_bar_table_chair}
    \end{minipage}
\end{center}

\begin{center}
    \begin{minipage}{0.98\linewidth}
        \centering
        \includegraphics[width=\linewidth]{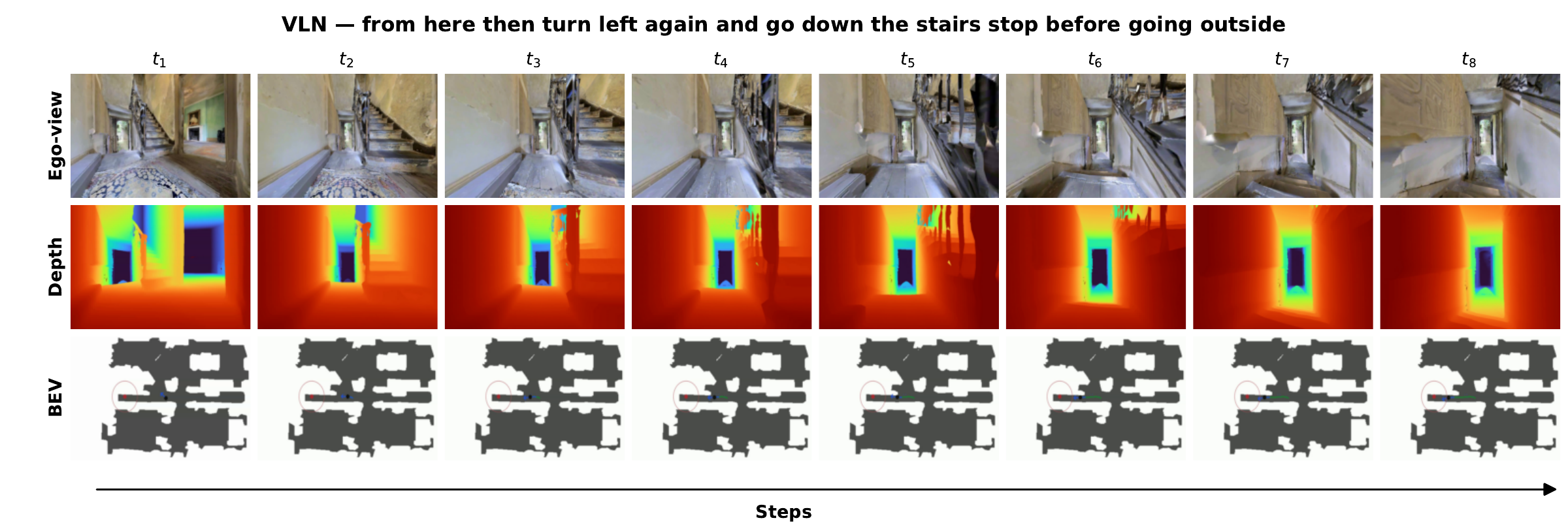}
        \captionof{figure}{\textbf{VLN simulation task, stairs before outside.} Qualitative visualization.}
        \label{fig:ablation_vln_stairs_outside}
    \end{minipage}
\end{center}

\begin{center}
    \begin{minipage}{0.98\linewidth}
        \centering
        \includegraphics[width=\linewidth]{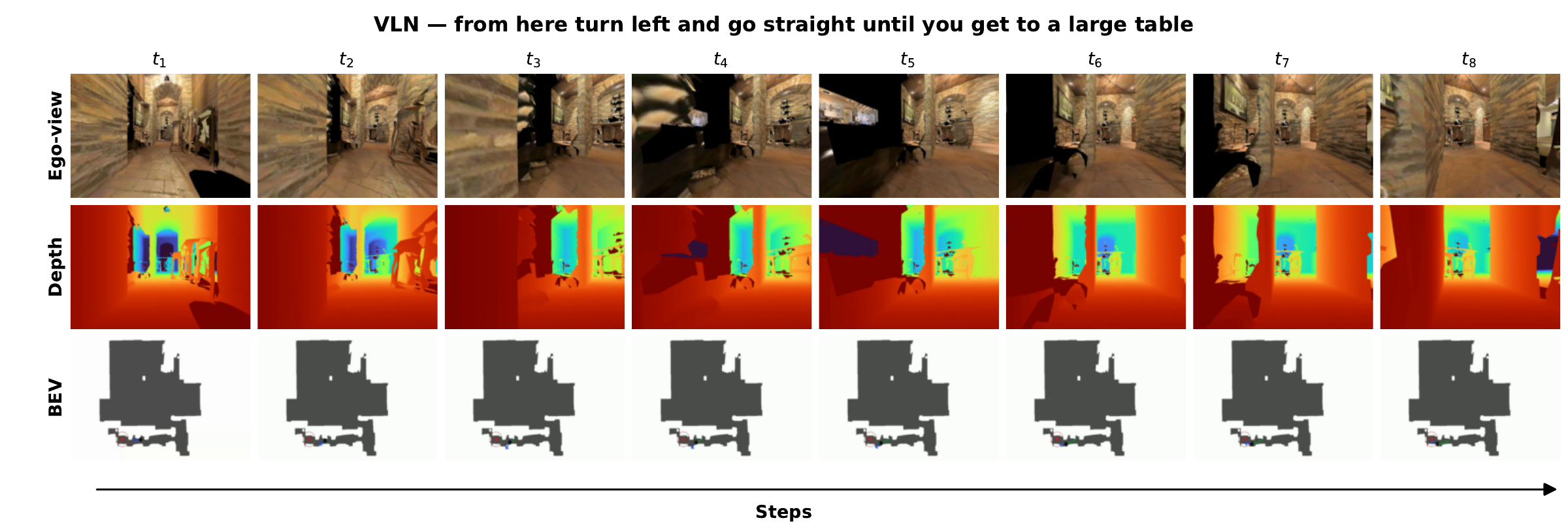}
        \captionof{figure}{\textbf{VLN simulation task, large table.} Qualitative visualization.}
        \label{fig:ablation_vln_large_table}
    \end{minipage}
\end{center}

\begin{center}
    \begin{minipage}{0.98\linewidth}
        \centering
        \includegraphics[width=\linewidth]{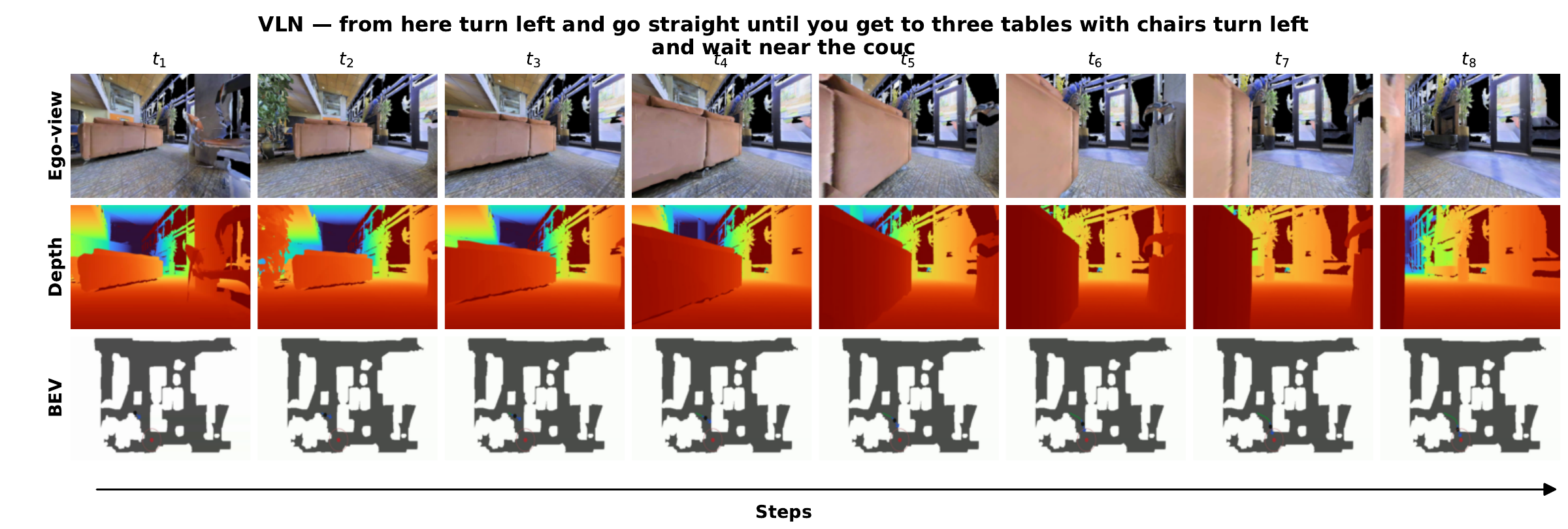}
        \captionof{figure}{\textbf{VLN simulation task, tables and chairs.} Qualitative visualization.}
        \label{fig:ablation_vln_tables_chairs}
    \end{minipage}
\end{center}

\begin{center}
    \begin{minipage}{0.98\linewidth}
        \centering
        \includegraphics[width=\linewidth]{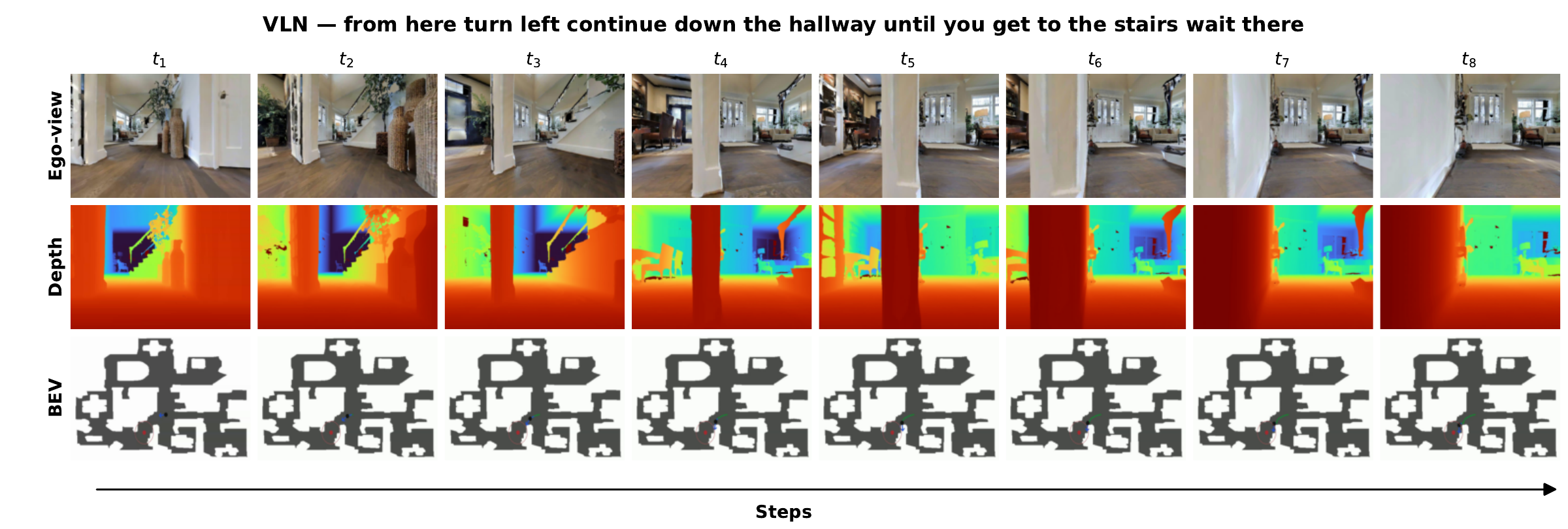}
        \captionof{figure}{\textbf{VLN simulation task, hallway to stairs.} Qualitative visualization.}
        \label{fig:ablation_vln_hallway_stairs}
    \end{minipage}
\end{center}

\begin{center}
    \begin{minipage}{0.98\linewidth}
        \centering
        \includegraphics[width=\linewidth]{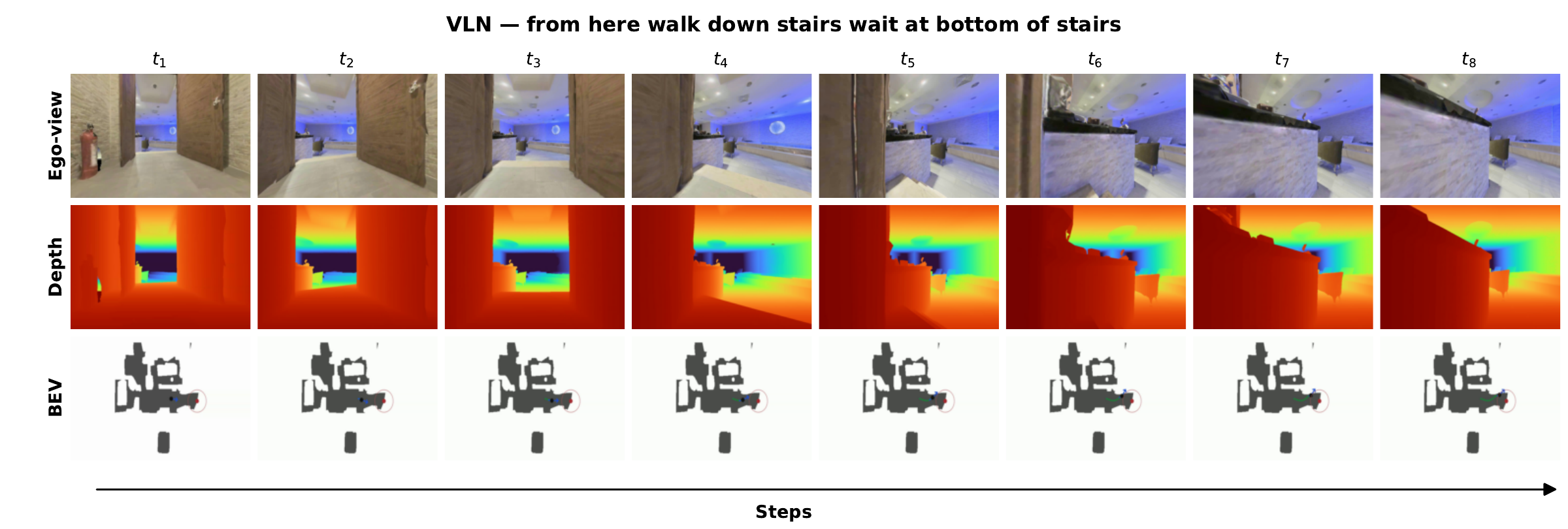}
        \captionof{figure}{\textbf{VLN simulation task, walk down stairs.} Qualitative visualization.}
        \label{fig:ablation_vln_walk_down_stairs}
    \end{minipage}
\end{center}

\begin{center}
    \begin{minipage}{0.98\linewidth}
        \centering
        \includegraphics[width=\linewidth]{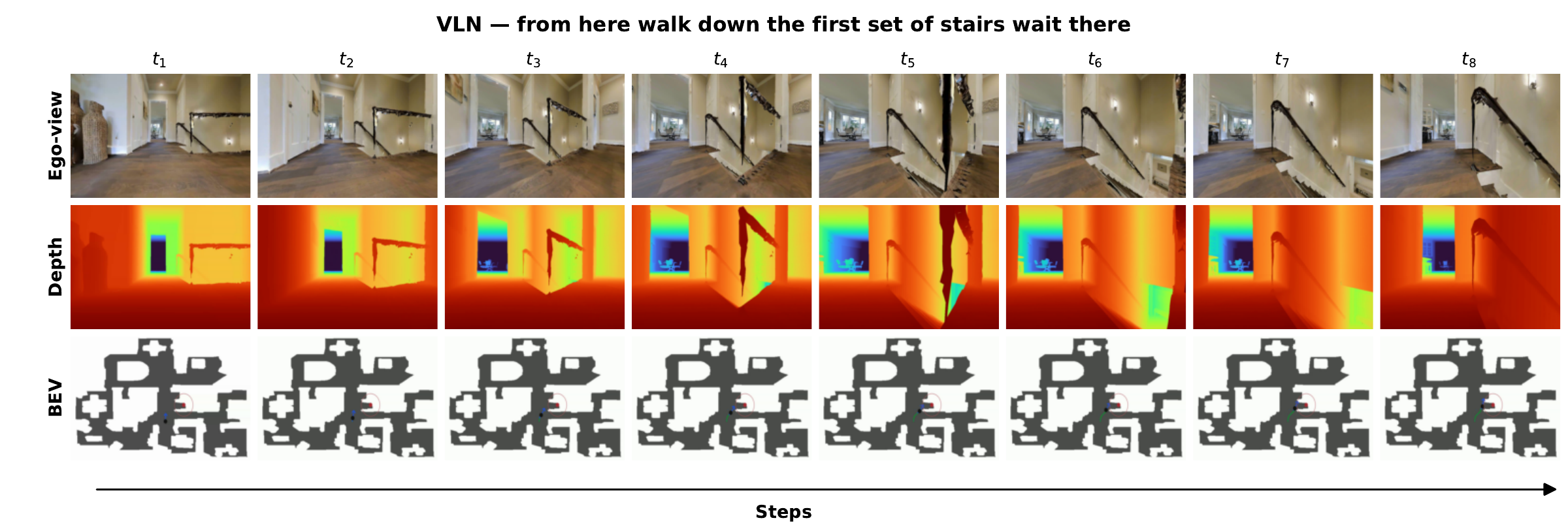}
        \captionof{figure}{\textbf{VLN simulation task, first set of stairs.} Qualitative visualization.}
        \label{fig:ablation_vln_first_stairs}
    \end{minipage}
\end{center}

\begin{center}
    \begin{minipage}{0.98\linewidth}
        \centering
        \includegraphics[width=\linewidth]{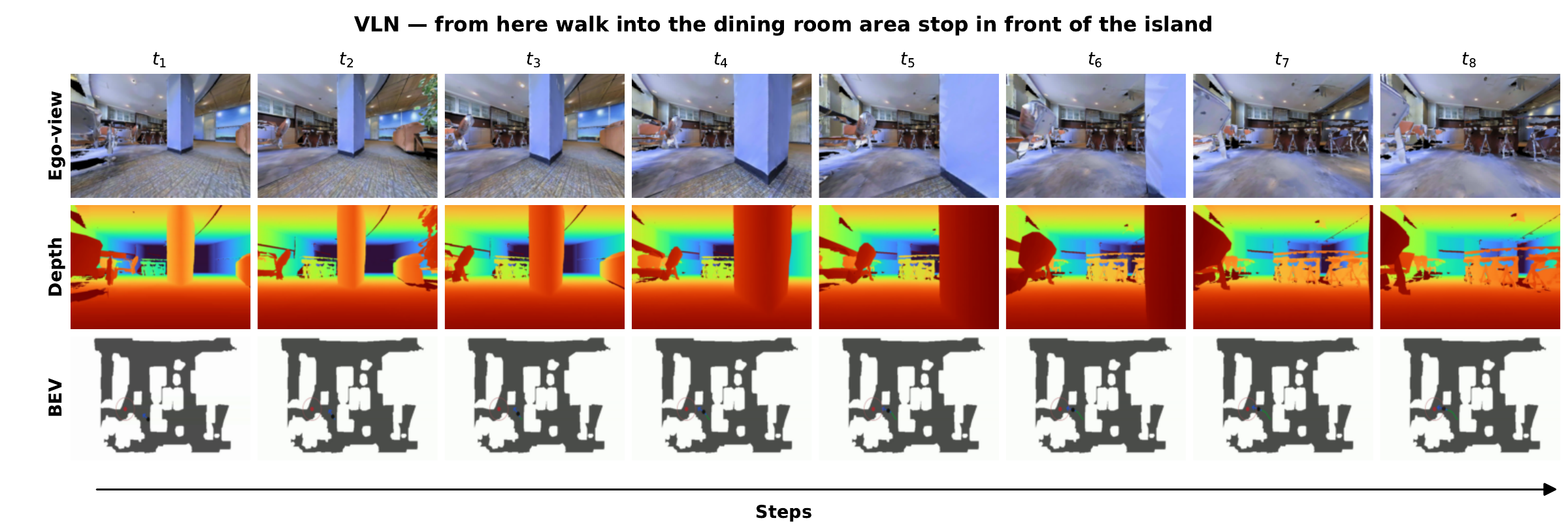}
        \captionof{figure}{\textbf{VLN simulation task, dining room island.} Qualitative visualization.}
        \label{fig:ablation_vln_dining_island}
    \end{minipage}
\end{center}

\begin{center}
    \begin{minipage}{0.98\linewidth}
        \centering
        \includegraphics[width=\linewidth]{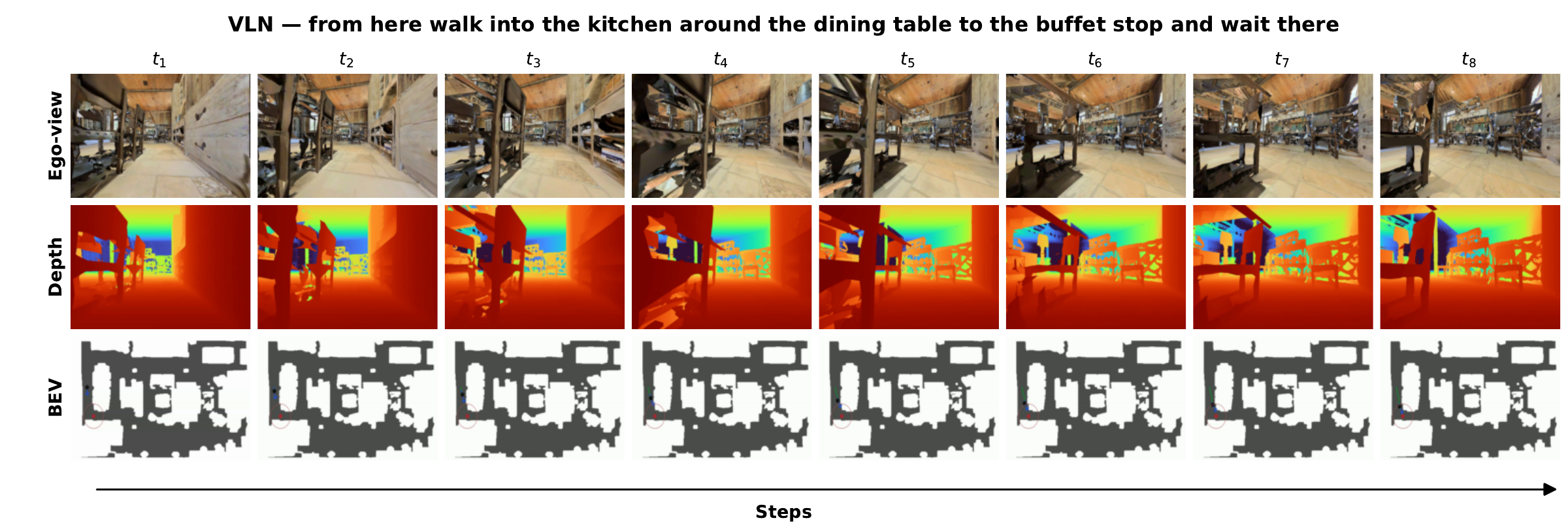}
        \captionof{figure}{\textbf{VLN simulation task, kitchen and buffet.} Qualitative visualization.}
        \label{fig:ablation_vln_kitchen_buffet}
    \end{minipage}
\end{center}

\begin{center}
    \begin{minipage}{0.98\linewidth}
        \centering
        \includegraphics[width=\linewidth]{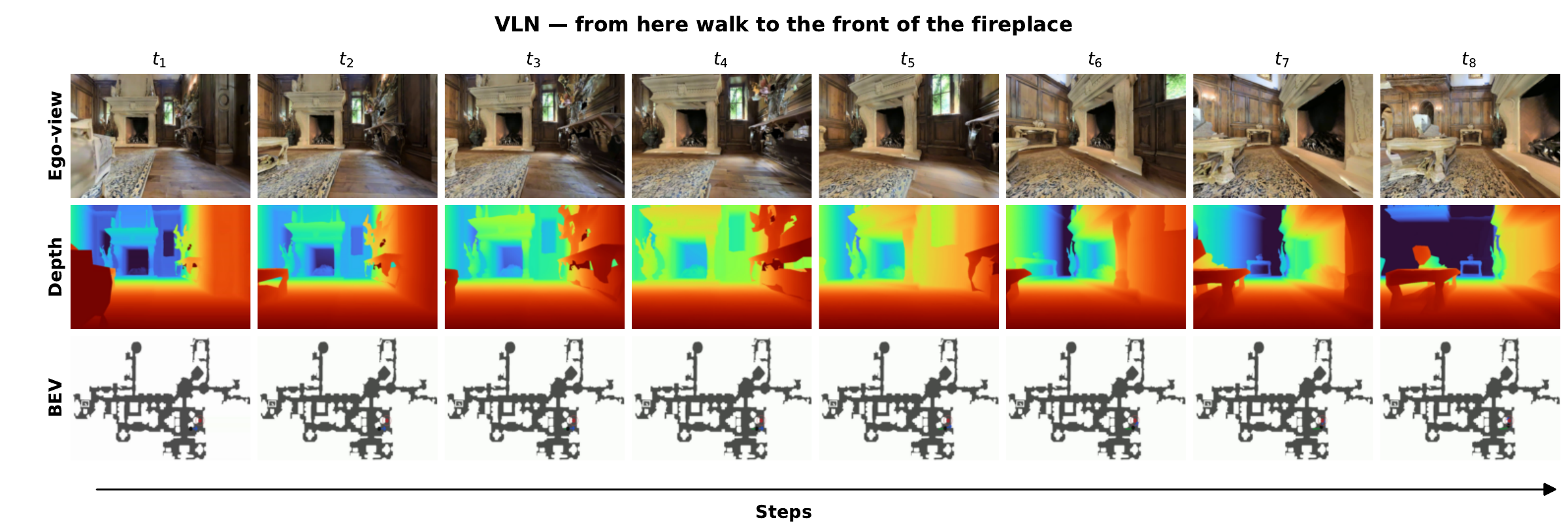}
        \captionof{figure}{\textbf{VLN simulation task, fireplace.} Qualitative visualization.}
        \label{fig:ablation_vln_fireplace}
    \end{minipage}
\end{center}

\begin{center}
    \begin{minipage}{0.98\linewidth}
        \centering
        \includegraphics[width=\linewidth]{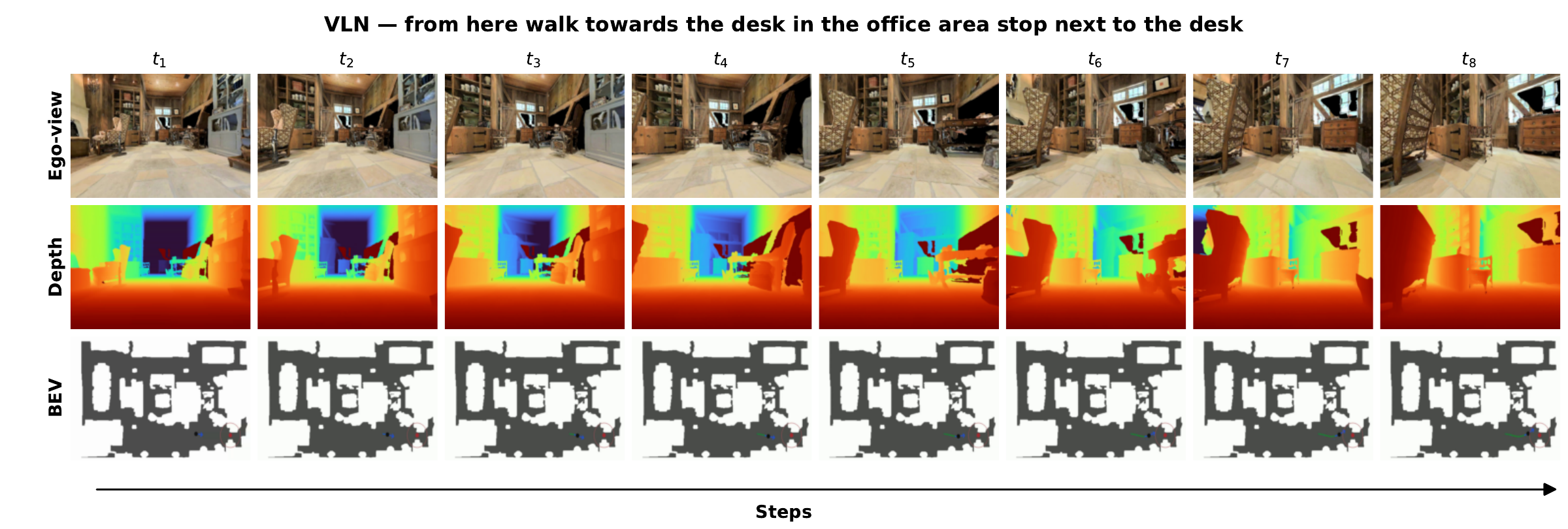}
        \captionof{figure}{\textbf{VLN simulation task, office desk.} Qualitative visualization.}
        \label{fig:ablation_vln_office_desk}
    \end{minipage}
\end{center}

\subsection{ObjNav Simulation Results}

\begin{center}
    \begin{minipage}{0.98\linewidth}
        \centering
        \includegraphics[width=\linewidth]{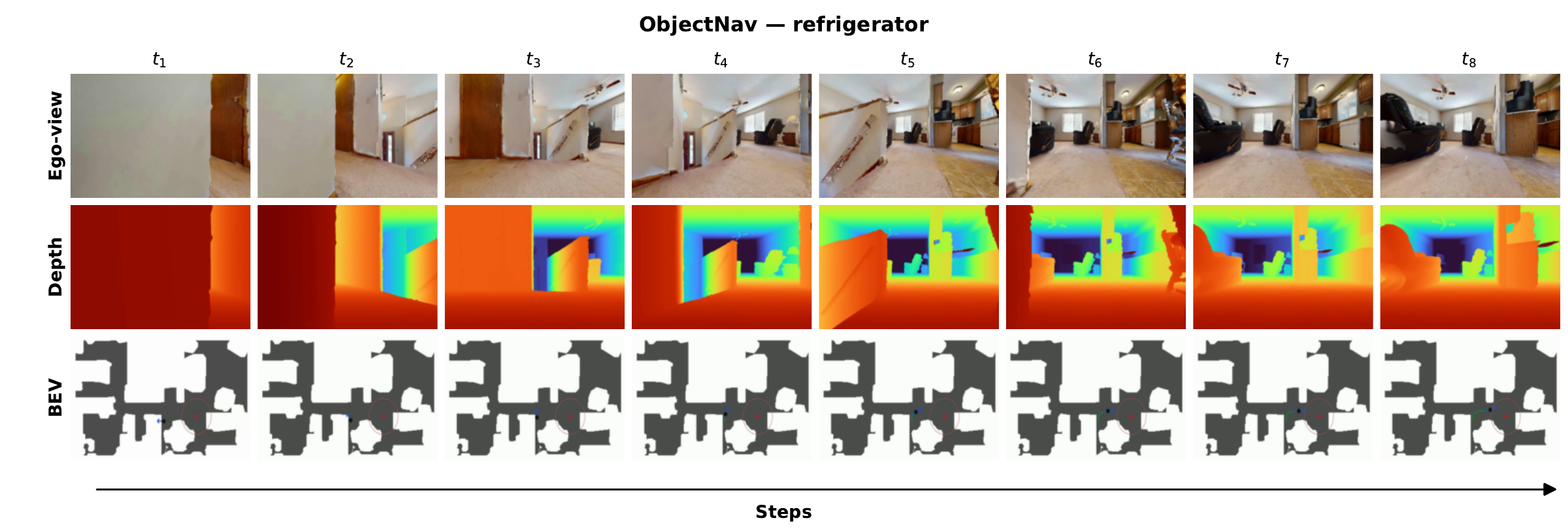}
        \captionof{figure}{\textbf{ObjNav simulation task, refrigerator.} Qualitative visualization.}
        \label{fig:ablation_platonic_refrigerator}
    \end{minipage}
\end{center}

\begin{center}
    \begin{minipage}{0.98\linewidth}
        \centering
        \includegraphics[width=\linewidth]{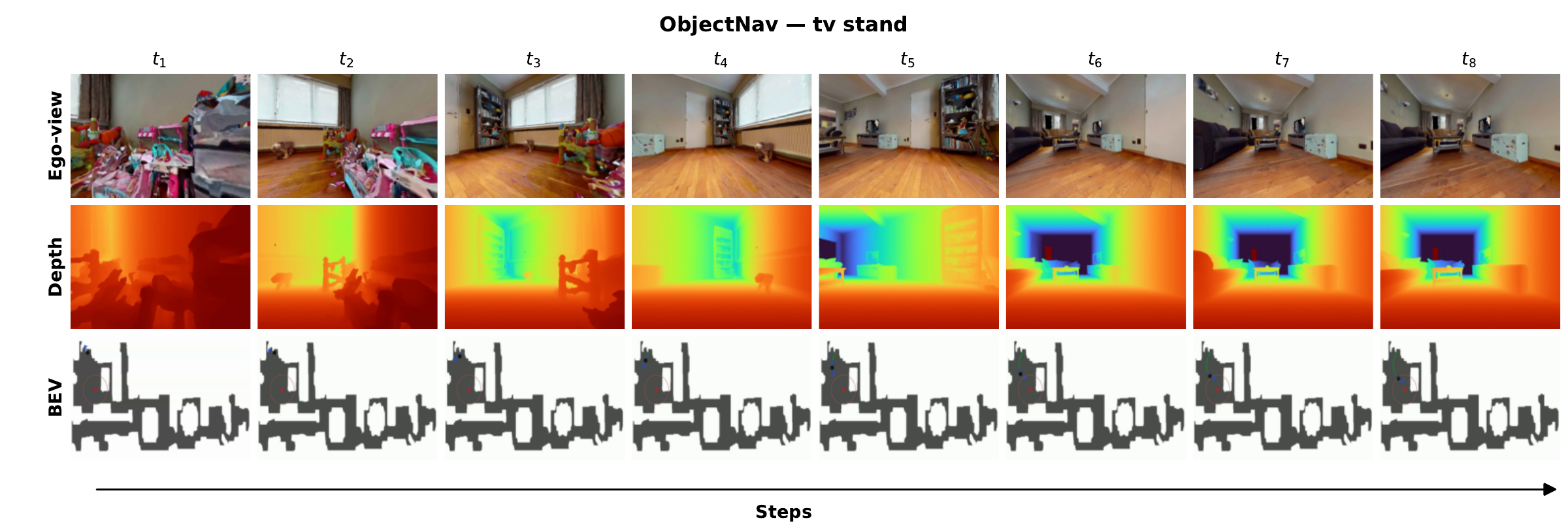}
        \captionof{figure}{\textbf{ObjNav simulation task, TV stand.} Qualitative visualization.}
        \label{fig:ablation_platonic_tv_stand}
    \end{minipage}
\end{center}

\begin{center}
    \begin{minipage}{0.98\linewidth}
        \centering
        \includegraphics[width=\linewidth]{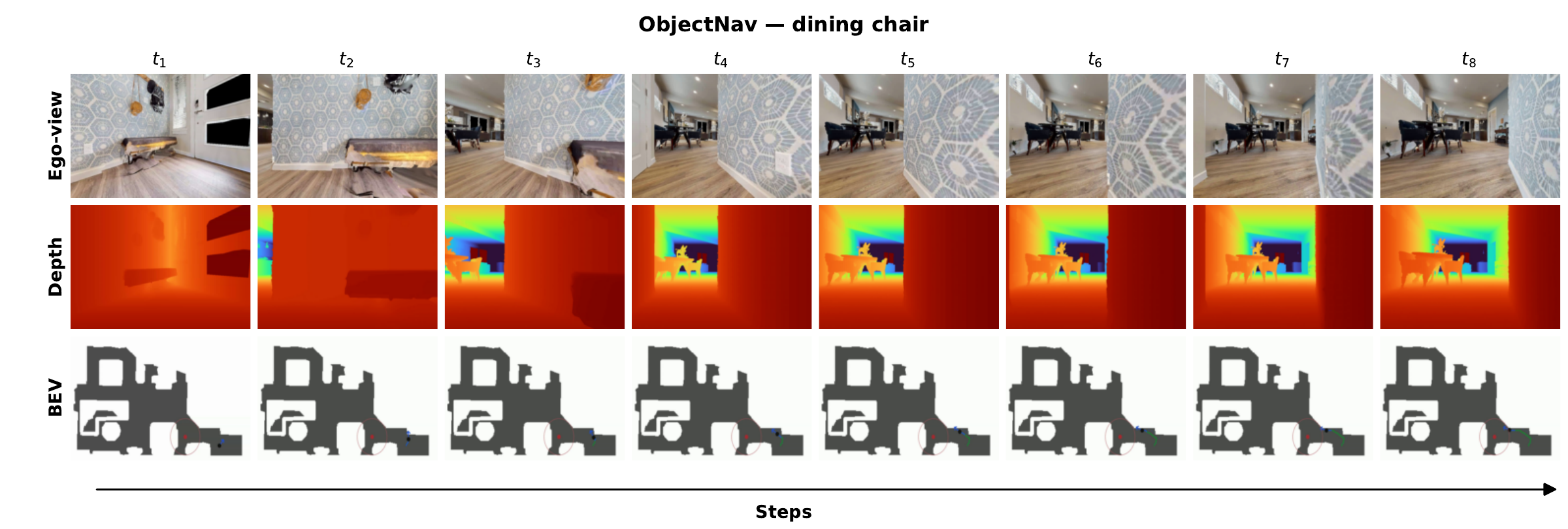}
        \captionof{figure}{\textbf{ObjNav simulation task, dining chair.} Qualitative visualization.}
        \label{fig:ablation_platonic_dining_chair}
    \end{minipage}
\end{center}

\begin{center}
    \begin{minipage}{0.98\linewidth}
        \centering
        \includegraphics[width=\linewidth]{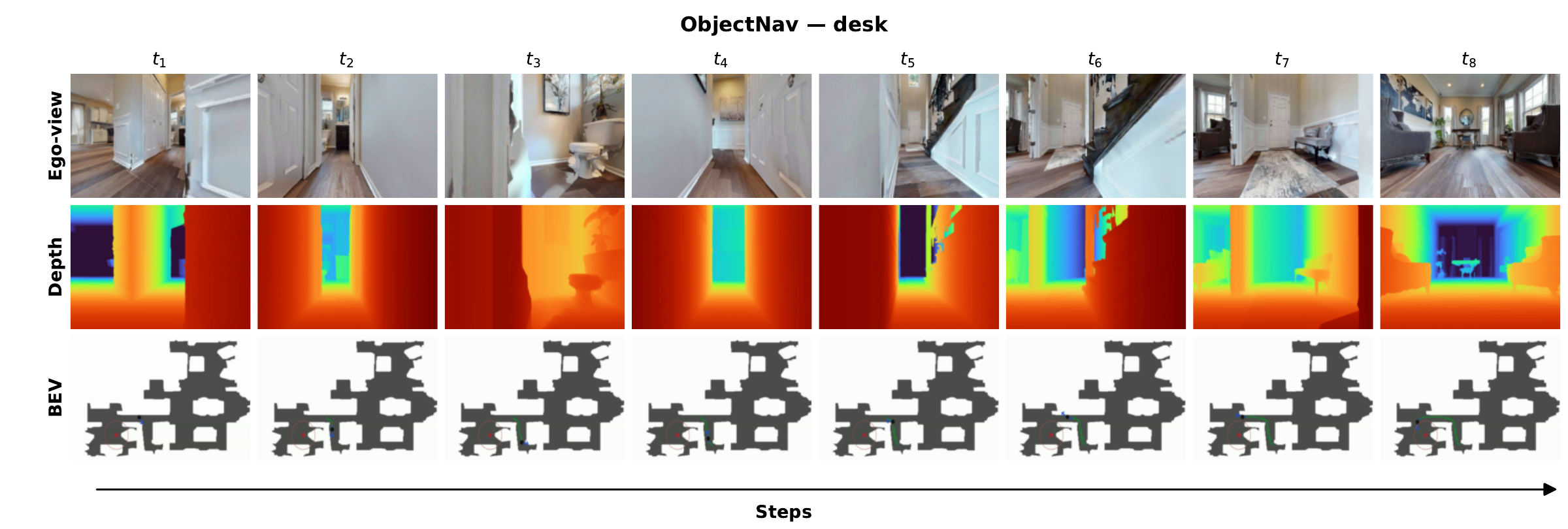}
        \captionof{figure}{\textbf{ObjNav simulation task, desk.} Qualitative visualization.}
        \label{fig:ablation_platonic_desk}
    \end{minipage}
\end{center}

\begin{center}
    \begin{minipage}{0.98\linewidth}
        \centering
        \includegraphics[width=\linewidth]{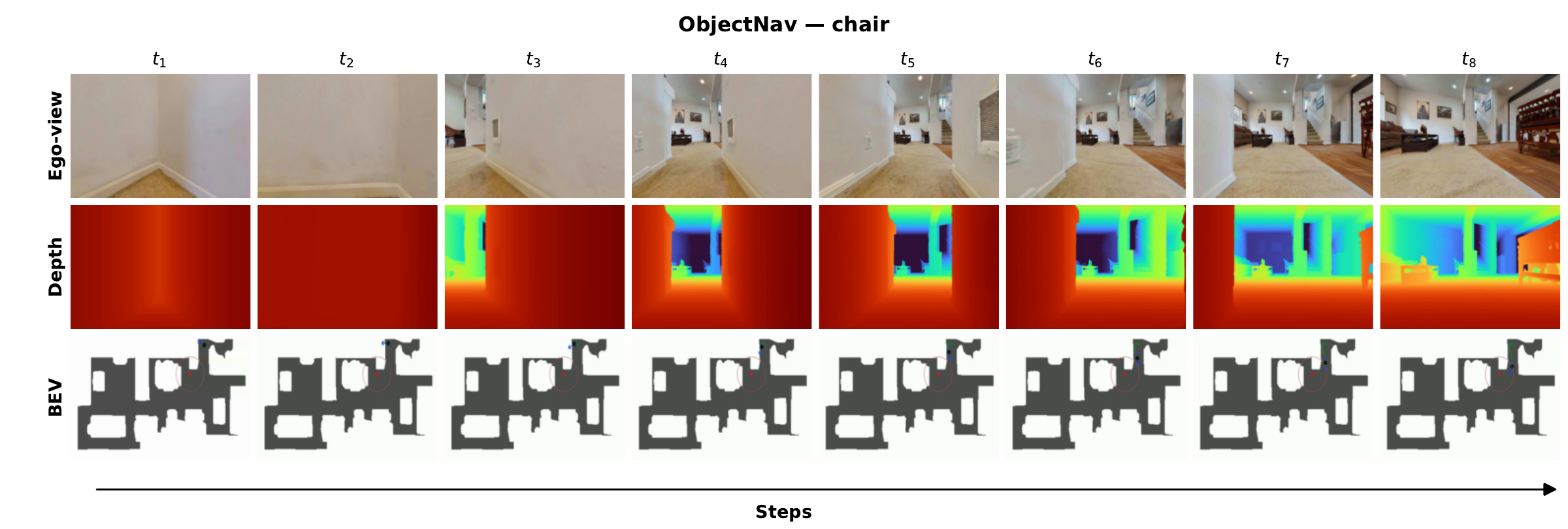}
        \captionof{figure}{\textbf{ObjNav simulation task, chair.} Qualitative visualization.}
        \label{fig:ablation_platonic_chair}
    \end{minipage}
\end{center}

\begin{center}
    \begin{minipage}{0.98\linewidth}
        \centering
        \includegraphics[width=\linewidth]{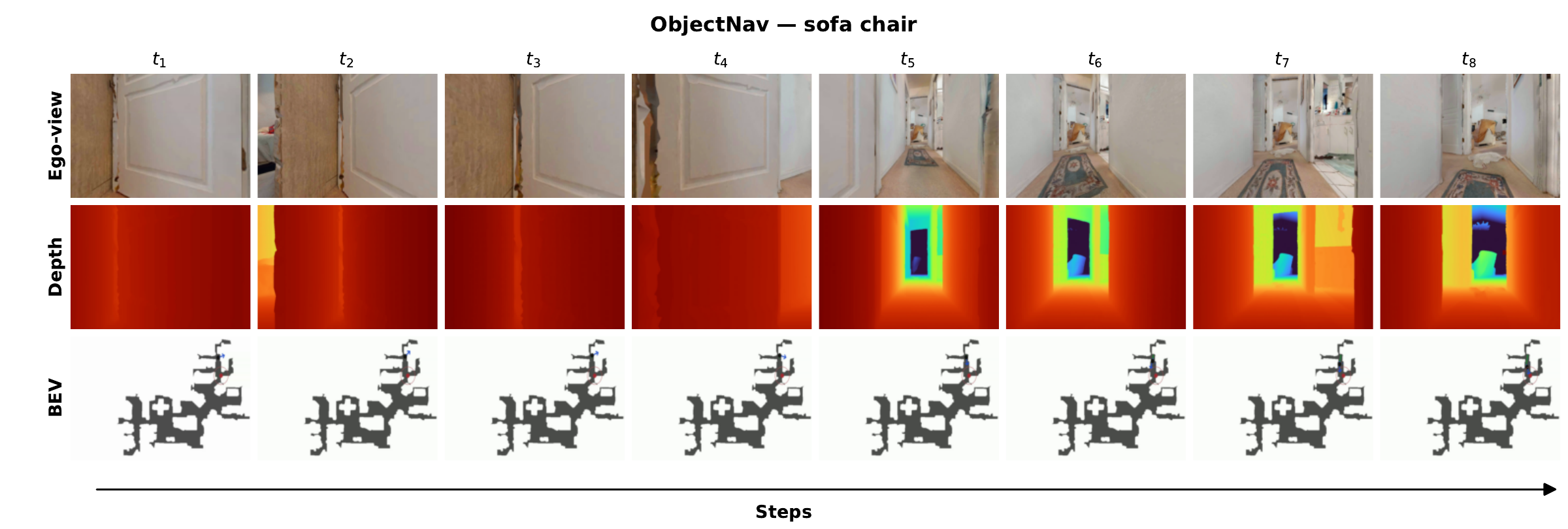}
        \captionof{figure}{\textbf{ObjNav simulation task, sofa chair.} Qualitative visualization.}
        \label{fig:ablation_platonic_sofa_chair}
    \end{minipage}
\end{center}

\begin{figure}[t]
    \centering
    \includegraphics[width=0.98\linewidth]{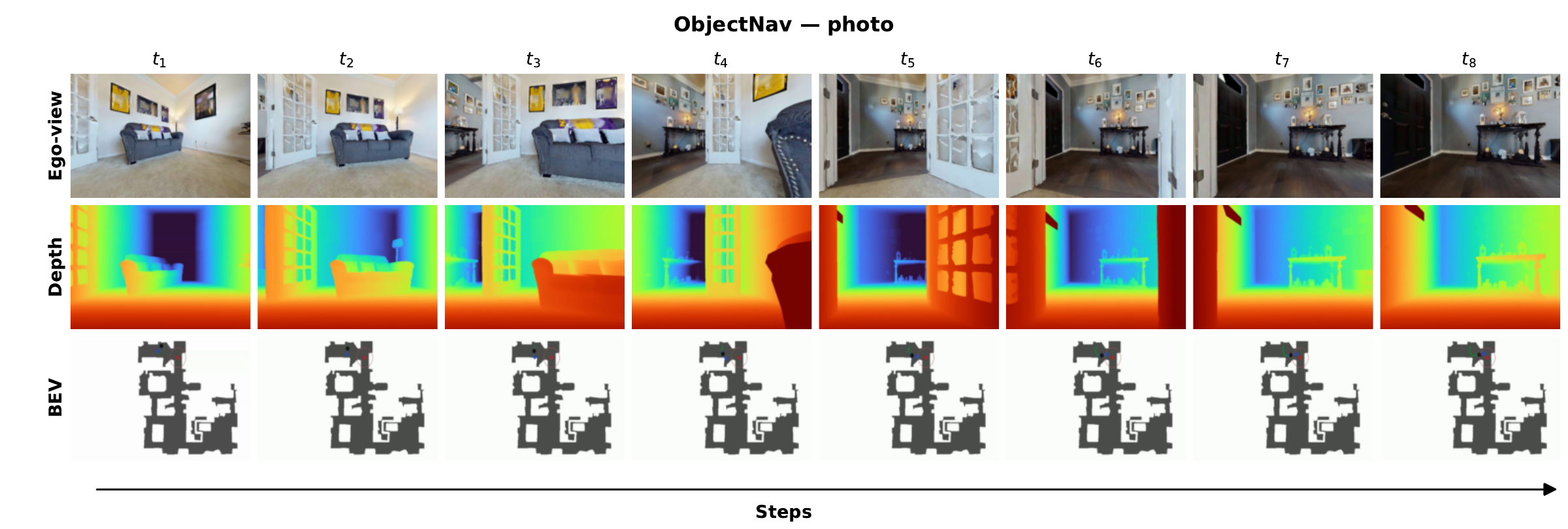}
    \captionof{figure}{\textbf{ObjNav simulation task, photo.} Qualitative visualization.}
    \label{fig:ablation_platonic_photo}
\end{figure}

\section{External Assets}
\label{sec:external-lic}
We list the existing assets used in PlatonicNav, together with their versions or identifiers and license terms.

\noindent\textbf{\href{https://github.com/matterport/habitat-matterport-3dresearch}{HM3D}}: v0.2; Matterport End User License Agreement for Academic Use of Model Data.

\noindent\textbf{\href{https://github.com/facebookresearch/habitat-matterport3d-dataset}{HM3D-IIN}}: HM3D Instance ImageNav v3; MIT code; HM3D-derived data under Matterport HM3D terms.

\noindent\textbf{\href{https://github.com/naokiyokoyama/ovon}{HM3D-OVON}}: official episodes; MIT-listed release; HM3D-derived data under Matterport HM3D terms.

\noindent\textbf{\href{https://niessner.github.io/Matterport/}{Matterport3D}}: v1; data under Matterport3D Terms of Use; code under MIT.

\noindent\textbf{\href{https://github.com/jacobkrantz/VLN-CE}{R2R-CE / VLN-CE}}: R2R-VLNCE v1-3; MIT code; Matterport3D-governed data.

\noindent\textbf{\href{https://github.com/MarSaKi/ETPNav}{ETPNav trajectories}}: official release; MIT code; Matterport3D-governed trajectory data.

\noindent\textbf{\href{https://github.com/facebookresearch/habitat-sim}{Habitat-Sim}}: v0.2.4 and v0.1.7; MIT license.

\noindent\textbf{\href{https://github.com/facebookresearch/habitat-lab}{Habitat-Lab}}: v0.2.4 and v0.1.7; MIT license.

\noindent\textbf{\href{https://github.com/oravus/object-rel-nav}{ObjectReact}}: used as an academic comparison baseline; no upstream license file is provided.

\noindent\textbf{\href{https://github.com/dominik-schnaus/itsamatch}{It's a Match}}: vendored implementation; MIT license.

\noindent\textbf{\href{https://github.com/facebookresearch/dinov3}{DINOv3 code}}: vendored implementation; DINOv3 License.

\noindent\textbf{\href{https://github.com/facebookresearch/sam2}{SAM2 code}}: vendored implementation; Apache-2.0 license.

\noindent\textbf{\href{https://huggingface.co/facebook/dinov3-vitb16-pretrain-lvd1689m}{DINOv3 ViT-B/16}}: LVD-1689M checkpoint; DINOv3 License.

\noindent\textbf{\href{https://github.com/facebookresearch/sam2}{SAM2 Hiera}}: official weights; Apache-2.0 license.

\noindent\textbf{\href{https://huggingface.co/sentence-transformers/gtr-t5-base}{GTR-T5-base}}: Sentence-Transformers checkpoint; Apache-2.0 license.

\section{Compute Resources}
\label{sec: compute-res}
All experiments were run on a SLURM cluster. Each PlatonicNav run uses a single NVIDIA H100 GPU with 4--8 CPU cores and up to 64\,GB RAM; auxiliary mapping and feature-caching jobs use a single NVIDIA L40 GPU. The four reported experiments together consumed roughly 100--150 H100-hours, with a comparable amount of additional compute spent on earlier method iterations not included in the paper.

\end{document}